\documentclass[journal]{IEEEtran}
\usepackage{comment}

\usepackage{tikz}
\usetikzlibrary{calc}

\tikzstyle{matrix of math nodes}=[%
  matrix of nodes,
  nodes={%
   execute at begin node=$,%
   execute at end node=$%
  }%
]

\title{\LARGE \bf
Saltation Matrices: \\The Essential Tool for Linearizing Hybrid Dynamical Systems
}

\author{Nathan J. Kong, J. Joe Payne, James Zhu, and Aaron M. Johnson%
\thanks{This material is based upon work supported by the U.S. Army Research Office under grant \#W911NF-19-1-0080 and the National Science Foundation under grants \#ECCS-1924723 and \#CMMI-1943900. The views and conclusions contained in this document are those of the authors and should not be interpreted as representing the official policies, either expressed or implied, of the Army Research Office, National Science Foundation, or the U.S. Government. The U.S. Government is authorized to reproduce and distribute reprints for Government purposes notwithstanding any copyright notation herein. }%
\thanks{Department of Mechanical Engineering, Carnegie Mellon University, Pittsburgh, PA. Corresponding author A.~M.~Johnson, {\tt\small amj1@cmu.edu}.}%
}

\usepackage{amsmath}
\usepackage{mathtools}
\usepackage{amssymb}
\usepackage{algorithm}
\usepackage{algpseudocode}
\usepackage{accents}
\usepackage{multirow}
\usepackage{cite}
\usepackage{subcaption}
\usepackage{empheq}
\usepackage{breqn}
\usepackage{xcolor}
\usepackage{amssymb}
\usepackage{pifont}
\usepackage[colorlinks,bookmarksopen,bookmarksnumbered,citecolor=black,urlcolor=red]{hyperref}
\newcommand{\cmark}{\ding{51}}%
\newcommand{\xmark}{\ding{55}}%
\definecolor{myRed}{rgb}{0.73333333333,0.333333333333,0.4}
\definecolor{myBlue}{rgb}{0,0.2666666666666,0.533333333333}
\definecolor{myYellow}{rgb}{0.8666666666,0.666666666666,0.2}
\definecolor{myGreen}{rgb}{0.2196, 0.34118,0.13725}
\definecolor{myGrey}{rgb}{0.46274509803,0.4431372549,0.4431372549}

\usepackage{graphicx}

\algdef{SE}[SUBALG]{Indent}{EndIndent}{}{\algorithmicend\ }%
\algtext*{Indent}
\algtext*{EndIndent}

\newcounter{theorem}

\newtheorem{definition}[theorem]{Definition}

\allowdisplaybreaks

\everymath{\displaystyle}

\newcommand{\pert}[1]{\widetilde{#1}}
\newcommand{\id}{I_{n\times n}}
\newcommand{\idm}{I_{m\times m}}
\newcommand{\hot}{\text{h.o.t.}}
\newcommand{\der}{\textbf{$\mathrm{D}$}}

\newcommand{\revise}[1]{{\color{black}#1}}

\newcommand\copyrighttext{%
 \textcopyright 2024 IEEE. Personal use of this material is permitted.
  Permission from IEEE must be obtained for all other uses, in any current or future
  media, including reprinting/republishing this material for advertising or promotional
  purposes, creating new collective works, for resale or redistribution to servers or
  lists, or reuse of any copyrighted component of this work in other works.
  DOI: \href{https://doi.org/10.1109/JPROC.2024.3440211}{10.1109/JPROC.2024.3440211}}
\newcommand\copyrightnotice{%
\begin{tikzpicture}[remember picture,overlay]
\node[anchor=south,yshift=5pt] at (current page.south) {\fbox{\parbox{\dimexpr\textwidth-\fboxsep-\fboxrule\relax}{  \footnotesize \copyrighttext}}};
\end{tikzpicture}%
}

\begin{document}
\bstctlcite{IEEEexample:BSTcontrol} 

\thispagestyle{empty}
\setcounter{page}{0}
\begin{figure*}[t!]
\centering
\large
This paper has been published in Proceedings of the IEEE.\\

DOI: \href{https://doi.org/10.1109/JPROC.2024.3440211}{10.1109/JPROC.2024.3440211}\\

IEEE Explore: \href{https://ieeexplore.ieee.org/document/10638633}{https://ieeexplore.ieee.org/document/10638633}\\

~\\

Please cite the paper as:\\

Nathan J. Kong, J. Joe Payne, James Zhu, and Aaron M. Johnson. ``Saltation matrices: The essential tool for linearizing hybrid dynamical systems,'' \emph{Proceedings of the IEEE}, vol. 112, no. 6, pp. 585–608, 2024.\\

~\\

~\\

\copyrighttext
\vspace{400px}
\end{figure*}
\maketitle
\copyrightnotice

\thispagestyle{empty}
\pagestyle{empty}
\begin{abstract}
    Hybrid dynamical systems, i.e.\ systems that have both continuous and discrete states, are ubiquitous in engineering, but are difficult to work with due to their discontinuous transitions. 
    For example, a robot leg is able to exert very little control effort while it is in the air compared to when it is on the ground. 
    When the leg hits the ground, the penetrating velocity instantaneously collapses to zero.
    These instantaneous changes in dynamics and discontinuities (or jumps) in state make standard smooth tools for planning, estimation, control, and learning difficult for hybrid systems.
    One of the key tools for accounting for these jumps is called the saltation matrix.
    The saltation matrix is the sensitivity update when a hybrid jump occurs and has been used in a variety of fields including robotics, power circuits, and computational neuroscience.
    This paper presents an intuitive derivation of the saltation matrix and discusses what it captures, where it has been used in the past, how it is used for linear and quadratic forms, how it is computed for rigid body systems with unilateral constraints, and some of the structural properties of the saltation matrix in these cases.
\end{abstract}

\section{Introduction}

Many interesting problems in engineering can be modeled as hybrid dynamical systems, meaning that they involve both continuous and discrete evolution in state \cite{Back_Guckenheimer_Myers_1993,LygerosJohansson2003,goebel2009hybrid, johnson2016hybrid}. 
These systems can be hybrid, e.g.\ due to physical contact, a result of digital logic circuits, or they can be triggered by control --  reacting to sensor feedback or switching control modes.
Meanwhile, most of the tools that exist for planning, estimation, control, and learning assume continuous (if not smooth) systems.
A common strategy to adapt tools that were designed for smooth systems to hybrid systems is to minimize the effect of discontinuities \cite{invariant_impact_control,dbhop} e.g.\ by slowing down to near zero velocity at the time of an impact event \cite{raibert1989dynamically}.
However, these strategies do not make use of the underlying dynamics of the system and only seek to mitigate them.
This may work out for certain fully actuated systems, but many hybrid systems of interest are underactuated and cannot always cancel out the discontinuous dynamics.

Rather than assuming continuous dynamics, we present tools that account for the effects of discrete events.
Often, discrete events are called ``jumps'' or ``resets'' that map state from one continuous domain to another.
The key to capturing hybrid events is to both model what occurs at the moment of reset and what happens \revise{to initial variations, whose trajectories we will refer to as perturbed trajectories,} that reset at different times.
One might think that analyzing the evolution of these variations simply requires linearization of the dynamics by taking the Jacobian of the reset map, but this only captures part of the story.
It is just as important to capture the variation that arises from changes in reset timing.
If the hybrid modes have different dynamics at the boundary, then trajectories that spend a different amount of time in each mode will result in changes in variation.

The \emph{saltation matrix}, sometimes referred to as the jump matrix, captures the total variation caused by both event timing and reset dynamics and is the key tool to understanding the evolution of trajectories near a hybrid event up to first order. 
The saltation matrix originally appeared in  \cite[Eq. 3.5]{AIZERMAN19581065}, where it was used to analyze the stability of periodic motions.
Other major works include \cite{filippov1988book,leine2004dynamics,Ivanov1998-ff}.
It provides essential information about event driven hybrid systems that can be used for stability analysis as well as for creating efficient estimation and control algorithms \cite{paper:kong-skf-2021,paper:kong-ilqr-2021,paper:kong-hybrid-2022,paper:payne-uncertainty-2022,paper:zhu-hybrid-2022,rijnen2015optimal,rijnen2019hybrid,saccon2014sensitivity}.
The word ``saltation'' directly translates to ``leap'' from Latin -- which closely matches to the ``jump'' name for the hybrid events -- and is also used to describe how sand particles ``leap'' along the ground when blown by wind in the desert \cite{owen1964saltation}.

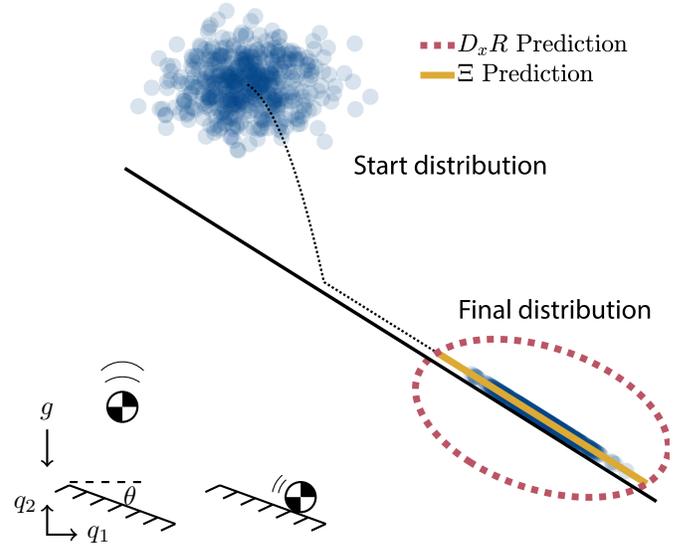
\begin{figure}[tb]
\centering
\begin{tikzpicture}[scale=1.0,thick]
    \node[anchor=south west,inner sep=0] at (0,0) {\includegraphics[width=0.4\textwidth]{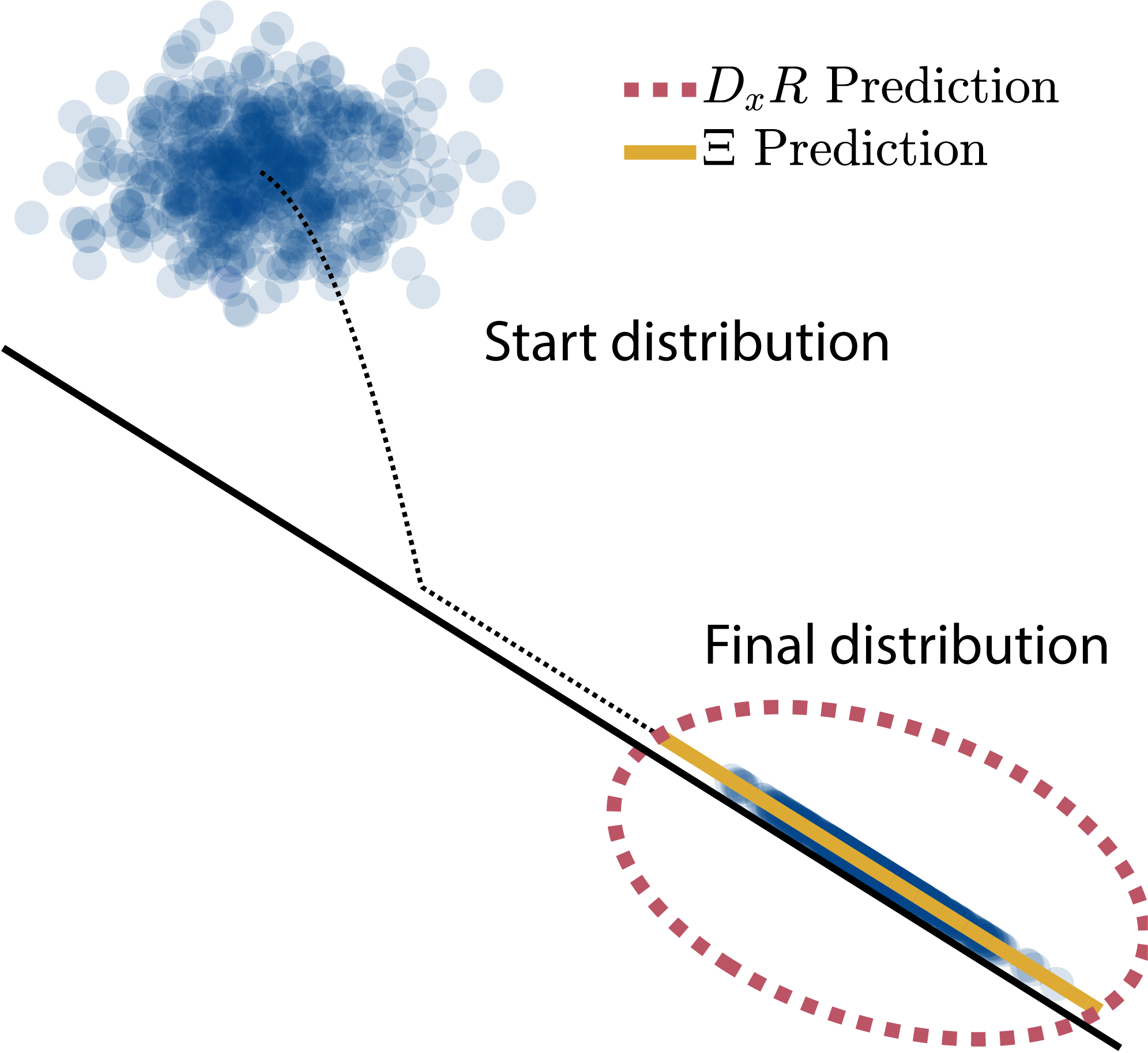}};
        \def\w{0.15}
    \def\ra{.2}
    \def\centerarc[#1](#2)(#3:#4:#5)
        { \draw[#1] ($(#2)+({#5*cos(#3)},{#5*sin(#3)})$) arc (#3:#4:#5); };
    
    \draw [<->] (-1,0)  node[left] {$q_2$} -- (-1,-.4) -- (-.6,-.4) node[right] {$q_1$};
    
    \draw [->] (-1,1) node[above] {$g$} -- (-1,.5);
    
    \begin{scope}[rotate={-20}]
    \draw [-] (-.75,0) -- (.75,0);
    \foreach \i in {-.75,-.5,...,.75}
        \draw (\i,0) -- (\i-\w,-\w);
    \end{scope}
    
    \begin{scope}[radius=\ra,shift={(0,1.3)}]
    \fill (0,0) -- ++(\ra,0) arc [start angle=0,end angle=90] -- ++(0,-\ra*2) arc [start angle=270, end angle=180];
    \draw (0,0) circle ;
    \centerarc[thin](0,0)(55:125:.4);
    \centerarc[thin](0,.2)(45:135:.4);
    \end{scope}
    
    \draw [dashed] (-.7,.3) -- (.3, .3);
    \node at (.1,.12) {$\theta$};
    
    
    \begin{scope}[shift={(2,0)}]
    
    \begin{scope}[rotate={-20}]
    \draw [-] (-.75,0) -- (.75,0);
    \foreach \i in {-.75,-.5,...,.75}
        \draw (\i,0) -- (\i-\w,-\w);
    \end{scope}
    
    \begin{scope}[radius=\ra,shift={(.37,\ra-.1)}]
    \fill (0,0) -- ++(\ra,0) arc [start angle=0,end angle=90] -- ++(0,-\ra*2) arc [start angle=270, end angle=180];
    \draw (0,0) circle ;
    \centerarc[thin](0,0)(130:170:.3);
    \centerarc[thin](0,0)(130:170:.38);
    \end{scope}
    \end{scope}

\end{tikzpicture}
        \caption[Example drop on a slanted surface with initial covariance]{Example drop on a slanted surface with initial covariance. The saltation matrix ($\Xi$) correctly estimates the end distribution's covariance where covariance in the direction of the constraint is eliminated. Using the incorrect update, only the Jacobian of the reset map  ($\der_x R$) leads to retaining belief in the direction of the constraint.
    }
    \label{fig:ball-drop-covariance}
\end{figure}

An illustrative example of how the saltation matrix can capture a common hybrid system, a rigid body with contact, is shown Fig. \ref{fig:ball-drop-covariance}. 
Here a distribution of balls is dropped on a slanted surface.
When each ball makes contact with the surface, a plastic impact law is applied which resets the system into a sliding mode on the surface by zeroing out the velocity into the surface.
For this system, the distribution starts out in the full 2D space and ends up constrained to the 1D surface after all balls have made impact.
However, since the reset map only changes the velocity of the ball, its Jacobian does not capture this change in the position variations. 
The saltation matrix captures this information and accurately predicts the resulting covariance by accounting for the difference in timing.
Sec. \ref{sec:saltcommon} in this tutorial shows that a similar trend is found for general rigid body contact systems.

\revise{The objective of this paper is twofold. First, we present a survey of the saltation matrix and its use in a number of areas from robotics to computational neuroscience, discussed in Sec. \ref{sec:survey}.
The rest of the paper, Sec. \ref{sec:definition}-\ref{sec:saltcommon}, presents a tutorial on the derivation of the saltation matrix and example computations for a simple but common class of hybrid systems relevant to robotics applications.
The ultimate goal of this work is to enable non-specialist controls and robotics engineers to understand the saltation matrix and be able to incorporate it into the analysis and design of algorithms. Specifically, this paper is organized as follows:}
\begin{itemize} 
    \item (Sec. \ref{sec:survey}) A literature survey of where the saltation matrix is being used in a variety of application areas.
    \item (Sec. \ref{sec:definition}) A tutorial on the definition of the saltation matrix (Sec.~\ref{sec:saltation}), its derivation (Sec.~\ref{sec:derivation}), and how it appears in linear (Sec.~\ref{subsec:linearforms}) and quadratic forms (Sec.~\ref{sec:quadraticforms}).
    \item (Sec. \ref{sec:salt_example}) An example showing the saltation matrix calculation for a simple contact system and a discussion of the properties of saltation matrices in various cases.
    \item (Sec. \ref{sec:saltcommon}) The calculation of saltation matrices for a common class of hybrid dynamical systems, rigid body dynamics with contact and friction, that unifies and extends prior analysis that has been scattered across different texts. This section provides more details on the properties of saltation matrices presented in (Sec. \ref{sec:salt_example}), including the eigenstructure of the saltation matrix for different cases.
\end{itemize}
In addition to providing a survey and tutorial for the saltation matrix, this paper also presents an alternate derivation of the saltation matrix using the chain rule (App. \ref{appendix:chainrule}), a derivation of the case in which the perturbed trajectory reaches a guard condition before the nominal trajectory (App. \ref{appendix:salt_derivation_early}), and derivations for how it is used to propagate covariances (App. \ref{appendix:covariance}) and to update the Riccati equations (App. \ref{appendix:Riccati}), all of which have not been presented previously.
\revise{These appendices are not crucial to the tutorial aspect of the paper, but provide further insight for experienced practitioners that desire a more rigorous discussion of the saltation matrix.}

\section{Survey of saltation matrix applications}
\label{sec:survey}
 \begin{figure*}
     \centering
     \includegraphics[width=\textwidth]{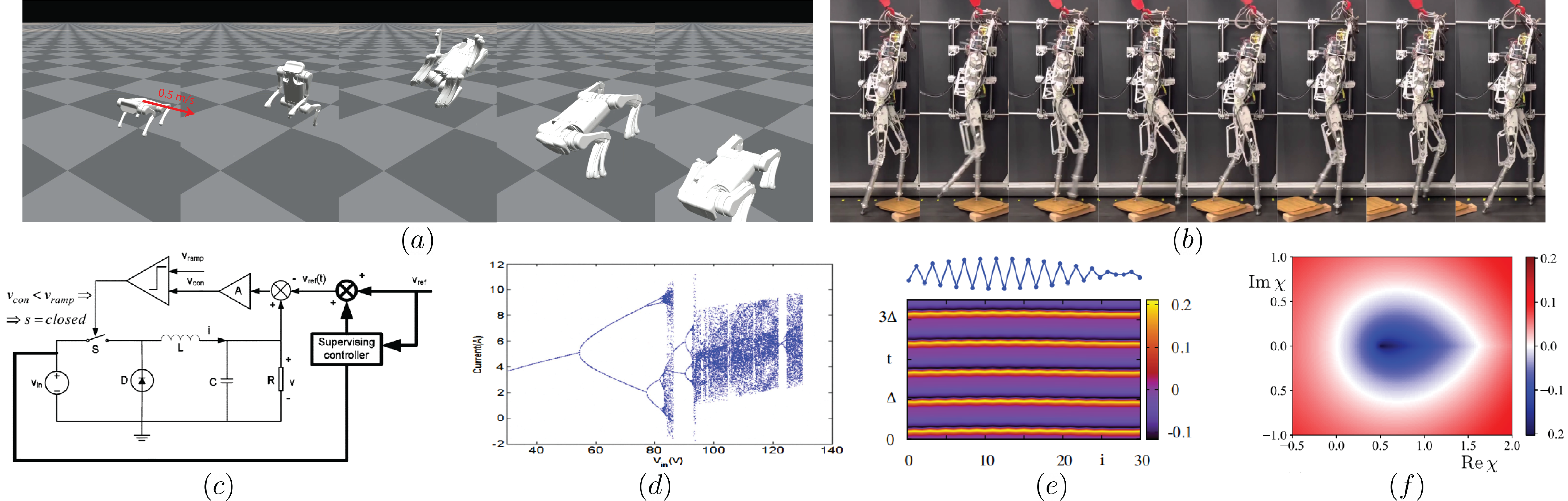}
     \caption{The saltation matrix has been used in many different fields, including the control of legged robots in tasks such as (a) a quadrupedal backflip \cite{paper:kong-hybrid-2022} and (b) robust bipdeal walking \cite{tucker2022bipedsaltation}; the analysis and control of power circuits such as (c) supervising control of a buck converter \cite{Giaouris2008Stability} and (d) the bifurcation behavior of DC drives \cite{okafor2010analysis}; and the modelling of neural activity in the brain such as (e) the stability analysis of a Wilson-Cowan neural mass model\cite{coombes2018networks} and (f) the modelling of synaptic filter behavior\cite{lai2018analysis}.}
     \label{fig:composite}
 \end{figure*}
The saltation matrix is a valuable tool for analysis and control in a wide variety of fields such as general bifurcations theory \cite{leine1999fold,leine2002discontinuous,leine2006bifurcation,di2008bifurcations,kowalczyk2011micro}, power circuits \cite{hiskens2000trajectory,ivanov2000stability,Maity2007ControlOB,Giaouris2008Stability,okafor2010analysis,bizzarri2011noise,giaouris2011complex,chakrabarty2012control,bizzarri2013extension,biggio2013reliable,mallik2020accurate}, rigid body systems \cite{banerjee2009invisible,revzen2015data,bizzarri2016necessary,suda2016why,jiang2017grazing,chawla2022stability}, chemical processing \cite{barton_discontinous_optimization_1998}, and hybrid neuron models \cite{bizzarri2013lyapunov,nobukawa2015chaotic,nobukawa2017chaotic,coombes2018networks,lai2018analysis,park2018infinitesimal}.
Fig. \ref{fig:composite} shows a few examples that demonstrate the usage of the saltation matrix in the legged robotics, power circuits, and neural modelling literature.


Often, the saltation matrix is used to assess the stability of hybrid dynamical systems, especially for periodic systems \cite{AIZERMAN19581065,leine2004dynamics,Ivanov1998-ff}.
The most popular method for analyzing stability of periodic hybrid systems is to analyze the fundamental matrix solution (as shown in Sec. \ref{subsec:linearforms}) which for periodic systems, is called the monodromy matrix\cite{lopez2004energy,bernardo2008piecewise,banerjee2009invisible,fevckan2010bifurcation,mandal2014new,jiang2017grazing,asahara2018stability,nicks2018clusters,chen2019calculating,dieci2021master,chawla2022stability}.
The monodromy matrix is heavily used in the circuits field specifically for determining local stability of switching power converters and determining if bifurcations occur  \cite{aroudi2015review,giaouris2006control,daho2008stability,elbkosh2008stability,giaouris2009application,okafor2010chaos,banerjee2011control,banerjee2011nonsmooth,bizzarri2011steady,morel2011application,bizzarri2012periodic,daho2012co,imrayed2012analysis,cortes2013design,giaouris2013stability,mandal2013dynamical,mandal2013new,bizzarri2014simulation,wu2014stability,maity2015modeling,abusorrah2017avoiding,el2017combined,mandal2017controldesign,el2018nonlinear,gkizas2018border,wu2018polynomial,munoz2019enhancing,chakrabarty2020dynamic,el2020piecewise,kuznyetsov2021calculation,munoz2021designing,morel2022open}.
See \cite{aroudi2015review} for an in depth review for analyzing the stability of switching mode power converters.
For more information on bifurcations in periodic systems, see \cite{muller1995calculation,bockman1991lyapunov,kunze2000lyapunov,ageno2005lyapunov} which discuss Lyapunov exponents (the rate of separation of infinitesimally close trajectories) for hybrid systems.

In \cite{paper:zhu-hybrid-2022}, the saltation matrix components of the monodromy matrix are used to analyze known robotic stabilizing phenomena such as paddle juggling and swing leg retraction.
The saltation matrix formulation reveals ``shape'' parameters, which are terms in the saltation matrix that are independent from the system's dynamics, but have an effect on the stability of the system.
These shape parameters can be optimized to generate stable open loop trajectories for complex hybrid systems that undergo periodic orbits.
\revise{A similar strategy was used in \cite{zhu2024convergent} to generate robust closed-loop trajectories for legged robots.}

A more restrictive but stronger form of stability analysis, known as contraction theory \cite{lohmiller1998contraction}, can be done by analyzing the convergence of neighboring trajectories through hybrid events \cite{burden2022contraction} -- where global asymptotic convergence is guaranteed if both the continuous-time flow and the saltation matrix are infinitesimally contractive.

Another version of stability was analyzed in \cite{CORNER201919,GALAN199917,DEBACKER1966168,barton_modeling_2002} as sensitivities to system parameters.
Adapted saltation conditions were used to characterize sensitivities across hybrid events.
These results were used to formulate and solve optimal design problems.

In addition to stability analysis, saltation matrices are also useful for generating controllers.
In optimal control, value functions are propagated along a trajectory to generate feedback controllers.
For linear time-varying LQR, sensitivity information about a trajectory is used to schedule optimal gains along that trajectory.
To implement optimal trajectory tracking for a hybrid system, \cite{saccon2014sensitivity} utilized the saltation matrix to update the sensitivity equation (as shown in Sec. \ref{sec:quadraticforms}).
Due to the sudden jump from the reset map, the optimal controller will also have a jump in the gain schedule, as first noted in \cite{schwerin1996parameter}. 
Other work further expanding and improving on \cite{saccon2014sensitivity} include \cite{rijnen2015optimal,rijnen2017control,rijnen2017reference,rijnen2019hybrid}.
A key concept from these works for tracking hybrid trajectories is ``reference spreading'' or ``reference extension'' which creates a new references by extending the pre-transition state through the guard and the post-transition state backwards in time.
If there is a mode mismatch, the correct reference extension is selected to track.

Using similar value function approximations and reference spreading, \cite{paper:kong-ilqr-2021} proposed a contact implicit trajectory optimization method by extending these ideas to iterative LQR (iLQR). This approach is able to generate both the nominal state trajectory and the feedback controller without having to specify the mode sequence in advance, as in \cite{von1999user,kelly2017introduction,schultz2009modeling,posa2016optimization}, or depend on complementarity constraints that are difficult to solve, as in \cite{posa2014direct,mordatch2012discovery}.
Recently, this hybrid iLQR has also been used as an online Model Predictive Controller (MPC) \cite{paper:kong-hybrid-2022}.

The saltation matrix has also been used to supplement the concept of hybrid zero dynamics to design robust controllers for bipedal robots.
In \cite{tucker2022bipedsaltation}, the norm of the saltation matrix is included in the optimal controller cost function to mitigate the divergent effects of impact.

State estimation uses sensitivity information in an analogous way, where the saltation matrix can be used to propagate covariance through a hybrid transition (Sec. \ref{sec:quadraticforms}).
The first paper to do this is \cite{biggio2014accurate}, which considers covariance propagation for power-spectral density calculation in circuits.
This covariance propagation law was also applied to Kalman filtering for hybrid dynamical systems \cite{paper:kong-skf-2021}.
This work has also been extended to covariance propagation with noisy guards and uncertainty in the reset map \cite{paper:payne-uncertainty-2022}. 
In \cite{gu2021invariant,gu2022invariant} hybrid dynamics are considered in an invariant extended Kalman filter for use on \revise{L}ie groups.
Using covariance propagation is powerful for state estimation because it efficiently maintains the belief of a distribution through hybrid events.
In \cite{paper:kong-skf-2021}, this ``Salted Kalman Filter'' runs with comparable accuracy to a hybrid particle filter, e.g.\ \cite{koutsoukos2002monitoring}, at a fraction of the computation time.
The main drawbacks are that it uses a Gaussian approximation, that the entire distribution is propagated instantaneously, and that it is not capable of keeping track of a split distribution that exists near a hybrid transition (whereas non-parametric filters like the particle filter can maintain a non-Gaussian and split distribution).


In cases where multiple guard conditions are met at the same time such as simultaneous leg touchdown, the hybrid event must be analyzed with another tool known as the Bouligand derivative (B-derivative)  \cite{Council_George2021-lf, pace2017piecewise, burden2016event,scholtes2012introduction,bernardo2008piecewise} as the saltation matrix only considers the effects of individual hybrid transition events. 
The B-derivative can be thought of as a set of composed saltation matrices which capture infinitesimal effects of differing transition sequences. 
The B-derivative has been used to analyze stability in systems with simultaneous impacts in \cite{pace2017piecewise}.

\revise{From this survey of saltation matrix applications, we see that there have been two primary strategies that use the saltation matrix to improved control and estimation for hybrid systems.
The first is leveraging the fact that the saltation matrix is the linear dynamics Jacobian for hybrid transitions, meaning that any algorithm that relies on a linearization of the dynamics, like the iterative linear quadratic regulator (iLQR) \cite{paper:kong-ilqr-2021} or the extended Kalman filter (EKF) \cite{paper:kong-skf-2021} can be naturally extended to hybrid systems with minimal additional algorithmic complexity.
The second usage appears in methods for controlling hybrid systems, where the saltation matrix provides information on stabilizing/destabilizing nature of a hybrid transition. \cite{paper:zhu-hybrid-2022,zhu2024convergent,tucker2022bipedsaltation} augment prior algorithms that either did not consider stabilizing effects of hybrid transitions (i.e.\ iLQR) or assumed these effects were dominated by continuous-time feedback control (i.e.\ hybrid zero dynamics).}

\begin{table}[tb]
\caption{Notation used and equation, definition, or section of introduction.}
\begin{tabular}{l|l}
\revise{$a_g$}& \revise{Acceleration due to gravity, Sec.~\ref{sec:example_dynamics}}\\
$A$&Linearized vector field matrix, \eqref{eq:deltaxdotminus}\\
$\mathbb{COV}$& Covariance \\
$\der_*$&Jacobian w.r.t $*$ \\
$\mathcal{D},D$&Hybrid domain, Def. \ref{def:hs}\\
$\mathbb{E}$&Expectation\\
$e$ & Coefficient of restitution, \eqref{eq::resetmapdagger}\\
$\mathcal{F},F$&Vector field, Def. \ref{def:hs}\\
$f$&Constraint force vector, \eqref{eq:blockdynamics}\\
$f_{\mathrm{n}},f_{\mathrm{t}}$&Normal and tangential constraint forces, \eqref{eq:coulomb}\\
$\mathcal{G},G,g$&Guard sets and guard function, Def. \ref{def:hs}\\
$\bar{g}$&Linearized guard function, \eqref{eq:guard_linearization}\\
$H$&Hamiltonian, \eqref{eq:hamiltonian}\\
$\hot$&Higher order terms, \revise{\eqref{eq:sensitivity}}\\
$I$&Identity matrix\\
\revise{$\mathrm{I},\mathrm{J},\hdots$}&Hybrid modes, Def. \ref{def:hs}\\
$i,j$ & Hybrid mode indexes, \eqref{eq:monodromydef}\\
$J$&Constraint Jacobian, Sec. \ref{sec:salt_example}\\
$\mathcal{J}$ & Set of discrete modes, Def. \ref{def:hs}\\
$L$ & Limit cycle, Sec. \ref{subsec:linearforms}\\
$\ell$&Loss function, \eqref{eq:opt_cost_to_go}\\
$M, C, N, \Upsilon$&Mass, Coriolis, nonlinear force, and input matrices, \eqref{eq:blockdynamics}\\
$M^\dagger$, $J^\dagger$, $\Lambda^\dagger$&Dagger elements for rigid body systems, \eqref{eq::dagger}\\
$m,n$ & Configuration \& state dimensions, Def. \ref{def:salt}, Sec.~\ref{sec:rbdynamics}\\
n, t & Normal or tangential direction constraints, Sec.~\ref{sec:rbdynamics}\\
$P$&Co-vector quadratic matrix, \eqref{eq::saltriccatiupdate}\\
$\mathcal{P}$&Poincaré map, Sec. \ref{subsec:linearforms}\\
$p$&Costate, Appendix \ref{appendix:Riccati}\\
$Q$&Penalty on state, Appendix \ref{appendix:Riccati}\\
$q$, $\dot{q}$, $\ddot{q}$&Configuration, velocity, and acceleration, Sec. \ref{sec:salt_example}\\
$\mathcal{R},R$&Reset map, Def. \ref{def:hs}\\
$\bar{R}$&Linearized reset map, \eqref{eq:reset_linearization}\\
$\mathbb{R}$&Set of real numbers\\
$S$ & Poincaré section, Sec. \ref{subsec:linearforms}\\
$\mathcal{T}*$& Tangent bundle over *\\
$T$&Time period, Sec. \ref{subsec:linearforms}\\
$t$&Time, Sec. \ref{sec:definition}\\
$\pert{t}$&Perturbed impact time, Sec. \ref{sec:derivation}\\
$u$&Control input, Def. \ref{def:hs}\\
$\mathrm{U},\mathrm{V}, \mathrm{S}, \mathrm{C}$&Rigid body modes, Secs. \ref{sec:salt_example}, \ref{sec:saltcommon}\\
$V$&Penalty on input, Appendix \ref{appendix:Riccati}\\
$v,\lambda$ & Eigenvector and eigenvalue, Sec. \ref{sec:exanalysis}\\
$X$&Random variable, Appendix \ref{appendix:covariance}\\
$x$&State, Def. \ref{def:hs}\\
$x^*$ & Fixed point, Sec. \ref{subsec:linearforms}\\
$\pert{x}$&Perturbed trajectory, Sec. \ref{sec:derivation}\\
$\delta x$&Perturbation, \eqref{eq:delta_pert_x_t}\\
$Z$&Additional terms, \eqref{eq::saltdifference1}\\

$\Gamma$& Set of discrete transitions, Def. \ref{def:hs}\\
$\Delta$&Discrete timestep, Sec. \ref{subsec:linearforms}\\
$\theta$&Angle of sloped surface, \eqref{eq:gus}\\
$\mu$ & Floquet exponent, \revise{Sec.~\ref{subsec:linearforms}}\\
$\mu_s,\mu_k$&Static and kinetic friction coefficient, \eqref{eq:coulomb}\\
$\Xi$&Saltation matrix, \eqref{eq:saltationmatrix}\\
$\rho$&Random variable mean, Appendix \ref{appendix:covariance}\\
$\Sigma$&Covariance, Appendix \ref{appendix:covariance}\\
$\sigma$ & Floquet multiplier, \revise{Sec.~\ref{subsec:linearforms}} \\
$\tau$&Time to impact map, \eqref{eq:combined_flow}\\

$\Phi$&Monodromy matrix, \eqref{eq:monodromydef}\\
$\phi$&Solutions of the flow, \eqref{eq:phii}\\
$\Omega$&Saltation block element, \eqref{eq:salt_block_us}\\
$0$&Zero matrix \\
$(*)^-, (*)^+$&Pre-impact and post-impact, \revise{Sec.~\ref{sec:saltation}}\\
\end{tabular}
\label{tab:notation}
\end{table}

\section{The saltation matrix and how to use it}
\label{sec:definition}
This section defines the saltation matrix and the broad class of hybrid systems where the saltation matrix applies (Sec. \ref{sec:saltation}), derives the expression of the saltation matrix using a geometric approach (Sec. \ref{sec:derivation}), and demonstrates the use of saltation matrices in linear (Sec. \ref{subsec:linearforms}) and quadratic forms (Sec. \ref{sec:quadraticforms}). Table~\ref{tab:notation} summarizes the notation used throughout the rest of the paper.

\subsection{Saltation matrix definition}
\label{sec:saltation}

While there are many definitions of hybrid dynamical systems, e.g.\ \cite{Back_Guckenheimer_Myers_1993,LygerosJohansson2003,goebel2009hybrid,johnson2016hybrid}, this treatment of the saltation matrix is based on the definition from \cite{paper:kong-ilqr-2021}.

\begin{definition} \label{def:hs}
    A $C^r$ \textbf{hybrid dynamical system}, for continuity class $r\in \mathbb{N}_{>0} \cup \{\infty,\omega \}$, is a tuple $\mathcal{H} := (\mathcal{J},{\mathnormal{\Gamma}},\mathcal{D},\mathcal{F},\mathcal{G},\mathcal{R})$ where the parts are defined as:
    \begin{enumerate}
        \item $\mathcal{J} := \{\mathrm{I},\mathrm{J},...\} \subset \mathbb{N}$ is the finite set of discrete \textbf{modes}.
        \item $\mathnormal{\Gamma} \subseteq \mathcal{J}\times\mathcal{J}$ is the set of discrete \textbf{transitions} forming a directed graph structure over $\mathcal{J}$.
        \item $\mathcal{D}:=\amalg_{\mathrm{I}\in\mathcal{J}}$ ${D}_{\mathrm{I}}$ is the collection of \textbf{domains}, where $D_{\mathrm{I}}$ is a $C^r$ manifold and the state $x\in D_{\mathrm{I}}$ while in mode~$\mathrm{I}$.
        \item $\mathcal{F}\revise{:\mathbb{R} \times \mathcal{D} \rightarrow \mathcal{TD}}$ is a collection of $C^r$ time-varying \textbf{vector fields}, $F_{\mathrm{I}}\revise{:= \mathcal{F}|_{D_I} }:  \mathbb{R}\times D_{\mathrm{I}} \to\mathcal{T}D_{\mathrm{I}}$,  \revise{for each $\mathrm{I}\in\mathcal{J}$}.
        \item $\mathcal{G}:=\amalg_{(\mathrm{I},\mathrm{J})\in\mathnormal{\Gamma}}$ $G_{(\mathrm{I},\mathrm{J})}(t)$ is the collection of \textbf{guard sets}, where $G_{(\mathrm{I},\mathrm{J})}(t)\subseteq D_{\mathrm{I}} $ for each $(\mathrm{I},\mathrm{J})\in \mathnormal{\Gamma}$ is defined as a regular sublevel set of a $C^r$ guard function, i.e.\ $G_{(\mathrm{I},\mathrm{J})}(t)= \{x \in D_{\mathrm{I}} |g_{(\mathrm{I},\mathrm{J})}(t,x)\leq0\}$ and $\der_x g_{(\mathrm{I},\mathrm{J})}(t,x) \neq 0 \:\:\forall \:\:g_{(\mathrm{I},\mathrm{J})}(t,x) = 0$.
        \item $\mathcal{R}:\mathbb{R}\times \mathcal{G}\rightarrow \mathcal{D}$ is a $C^r$ map called the \textbf{reset} that restricts as $R_{(\mathrm{I},\mathrm{J})}:=\mathcal{R}|_{G_{(\mathrm{I},\mathrm{J})(t)}}:G_{(\mathrm{I},\mathrm{J})}(t)\rightarrow D_{\mathrm{J}}$ for each $(\mathrm{I},\mathrm{J})\in \mathnormal{\Gamma}$.
    \end{enumerate}
    Note that this definition incorporates the \textbf{control input} $u(t,x)$ into the dynamics $\mathcal{F}$ as $\mathcal{F}(t,x,u(t,x))$\revise{, which we simplify as} $\mathcal{F}(t,x)$ \revise{going forward}.
    \end{definition}

\begin{figure*}[t]
    \centering
    \includegraphics[width = 0.75\textwidth]{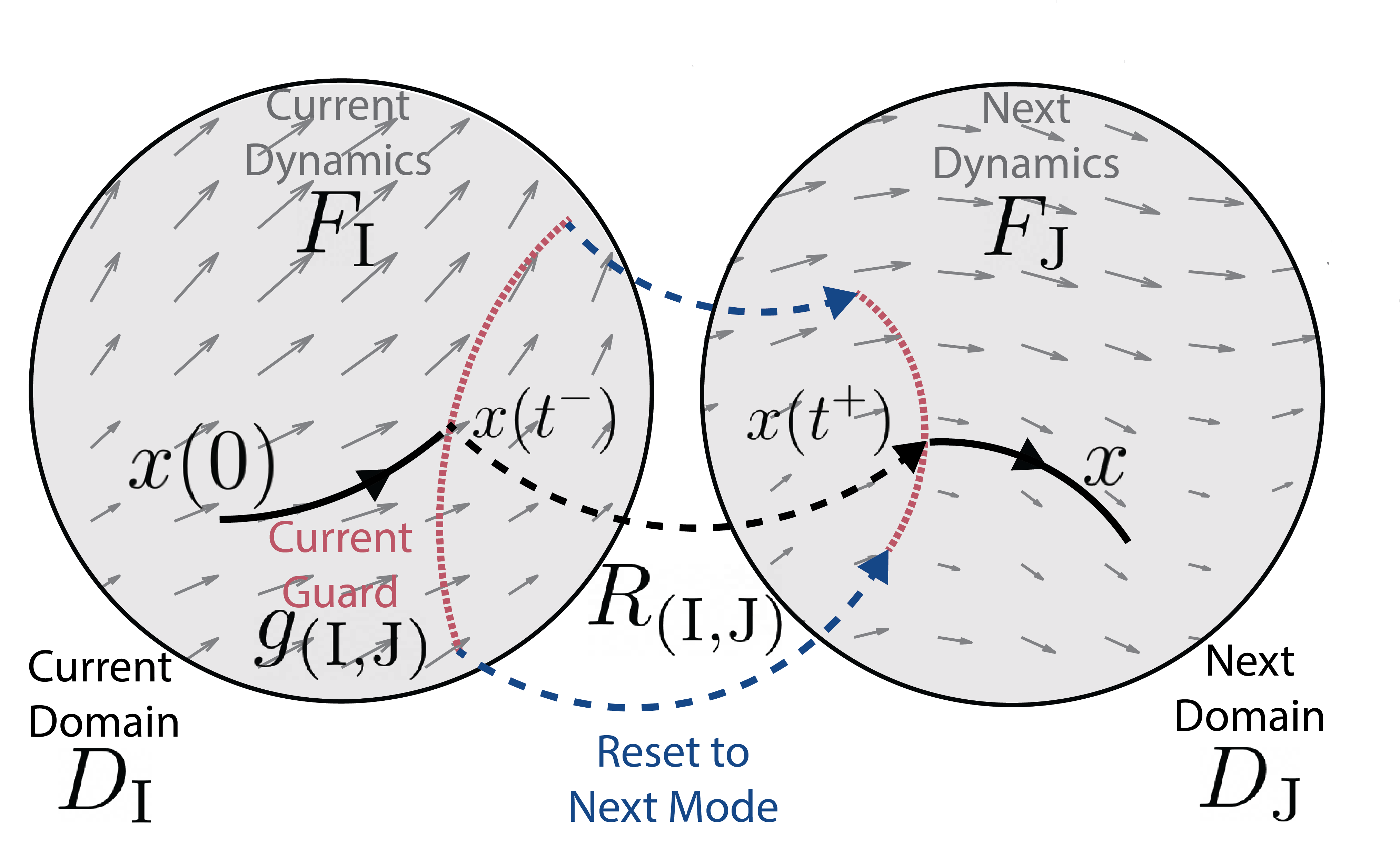}
    \caption[Example 2 mode hybrid system]{An example 2 mode hybrid system where the domains are shown in black circles $D$, the dynamics are shown with gray arrows $F$, the guard for the current domain is shown in red \revise{dotted} $g$, and the reset from the current mode to the next mode is shown in \revise{dashed lines} $R$.}
    \label{fig:hybrid_systems_diagram}
\end{figure*}

Fig.~\ref{fig:hybrid_systems_diagram} shows an example hybrid system with a hybrid execution consisting of a starting point $x(0)$ in ${D}_{\mathrm{I}}$ flowing with dynamics ${F}_{\mathrm{I}}$ and reaching the guard condition $g_{(\mathrm{I},\mathrm{J})}(t,x) = 0$ 
 at time $t$, applying the reset map $R_{(\mathrm{I},\mathrm{J})}(t,x)$ resetting into ${D}_{\mathrm{J}}$ and then flowing with the new dynamics ${F}_{\mathrm{J}}$.
 Denote $t^-$ as the instant before \revise{a hybrid event occurs while the system is still in domain $\mathrm{I}$}, $t^+$ the instant after the reset map is applied \revise{following the hybrid event where the system has transitioned into domain $\mathrm{J}$}, and $x(t^\pm) = x^\pm$ the limiting value of the signal $x$ from the left $(-)$ or right $(+)$.

    
The goal in this paper is to understand how \revise{variations} about a nominal trajectory evolve over time.
For smooth systems, it is well known that \revise{variations} about a nominal trajectory, \revise{$\delta x$}, can be approximated to first order using the derivative of the dynamics $F(t,x)$ with respect to state, \revise{$\der_x$}:
\begin{align}
    \revise{\frac{d}{dt}\delta x(t)} = \der_x F(t,x) \delta x \revise{+\hot} \label{eq:sensitivity}
\end{align}
\revise{where $\hot$ represents higher order terms.}
Hybrid systems with time triggered reset maps can be similarly analyzed using the Jacobian of the reset map, $\delta x^+ = \der_x R(t,x) \delta x^-$.
However, the Jacobian of the reset map does not account for differences that are introduced from time-to-impact variations in systems with event driven resets, where the differences in dynamics in the two hybrid modes must be considered.
The saltation matrix, e.g.\ \cite[Eq.~3.5]{AIZERMAN19581065}, \cite[Pg. 118 Eq. 6]{filippov1988book}, or \cite[Eq.~7.65]{leine2004dynamics}, accounts for these terms to capture how \revise{variations} are mapped through event-driven hybrid transitions to the first order.
From here on, the term hybrid transition/system refers to this event-driven class.

    \revise{For notational simplicity, the following shorthands} are made for the terms in the saltation matrix:
    \begin{align}
        F^-_{\mathrm{I}} &:= F_{\mathrm{I}}(t^-,x(t^-))\label{eq:F_I_minus}\\
        F^+_{\mathrm{J}} &:= F_{\mathrm{J}}(t^+,x(t^+))\label{eq:F_J_plus}\\
        x(t^+) &:= {R}_{(\mathrm{I},\mathrm{J})}(t^-,x(t^-))\\
        \der_xR^- &:= \der_x{R}_{(\mathrm{I},\mathrm{J})}(t^-,x(t^-))\label{eq:R_IJ_minus}\\ 
        \der_tR^- &:= \der_t{R}_{(\mathrm{I},\mathrm{J})}(t^-,x(t^-))\label{eq:D_t_R}\\
        \der_xg^- &:= \der_x{g}_{(\mathrm{I},\mathrm{J})}(t^-,x(t^-))\label{eq:g_IJ_minus}\\ 
        \der_tg^- &:= \der_t{g}_{(\mathrm{I},\mathrm{J})}(t^-,x(t^-))\label{eq:D_t_g}
    \end{align}
    Note that $\der_t$ in \eqref{eq:D_t_R} and \eqref{eq:D_t_g} refers to the derivative with respect to the first coordinate (and not the time dependence of $x$, which is captured by other terms). \revise{Now, we can define the saltation matrix as follows.}

\begin{definition} \label{def:salt}
    The \textbf{saltation matrix} for transition from mode I to mode J is the first order approximation of the variational update at hybrid transitions from mode I to J, defined as
\begin{empheq}[box=\fbox]{align}
        \Xi_{(\mathrm{I},\mathrm{J})} := \der_x R^-+\frac{\left(F^+_{\mathrm{J}}-\der_xR^- F^-_{\mathrm{I}} - \der_tR^-\right)   \der_x g^-}{\der_t g^- +\der_x g^-  F^-_{\mathrm{I}}} \label{eq:saltationmatrix}
    \end{empheq}
\revise{In the saltation matrix, the first term, $\der_xR^-$, captures the variations due to the reset map being applied at different states. The second term accounts for the variations caused by a trajectory being subject to differing dynamics for a small amount of time due to the displacement, which will be discussed in detail in Sec. \ref{sec:derivation}.}

    Note that the matrix multiplication in \eqref{eq:saltationmatrix} results in an outer-product between the terms in the parentheses and $\der_xg^-$ to get a rank-1 correction to the Jacobian of the reset map.
    The saltation matrix is an $n_{\mathrm{J}} \times n_{\mathrm{I}}$ matrix, where $n_{\mathrm{I}}$ is the dimension of the states in domain $D_{\mathrm{I}}$ and $n_{\mathrm{J}}$ is the dimension of the states in domain $D_{\mathrm{J}}$.

    %
    The saltation matrix maps \revise{variations} to the first order from pre-transition $\delta x(t^-)$ to post-transition $\delta x(t^+)$ as
    \begin{empheq}[box=\fbox]{align}
    \delta x(t^+) = \Xi_{(\mathrm{I},\mathrm{J})} \delta x(t^-) + \hot
    \label{eq::saltperturbation}
    \end{empheq}
    %
   
    \end{definition}
    
The saltation matrix in \eqref{eq:saltationmatrix} is \revise{well defined} when the following assumptions are true:
\begin{enumerate}
    \item Guards and resets are differentiable
    \item Trajectories must be transverse to the guard at an event: 
\begin{align}
 &\frac{d}{dt} g_{(\mathrm{I},\mathrm{J})}(t,x(t))= \der_tg^-+\der_x g^- F_{\mathrm{I}}^- < 0\label{asm:transverse}
\end{align}
\end{enumerate}

\revise{In addition, it is often taken that trajectories cannot undergo an infinite number of resets in finite time (no Zeno) in order to ensure trajectories can be analyzed without needing to determine the behavior in limit conditions.}

The saltation matrix relies on differentiating the guards and reset maps so they must be differentiable. 
Transversality ensures that neighboring trajectories impact the same guard unless the impact point lies on any other guard surface, in which case the Bouligand derivative is the appropriate analysis tool \cite{Council_George2021-lf, pace2017piecewise, burden2016event,scholtes2012introduction,bernardo2008piecewise}.
Transversality also ensures the denominator in \eqref{eq:saltationmatrix} does not approach zero.

\revise{These assumptions also indicate the main limitations of the saltation matrix.
On top of the limitations inherent to the linearization of nonlinear systems, the saltation matrix assumes that all neighboring trajectories undergo the same transition sequence as the nominal trajectory.
This is unable to capture situations where the nominal trajectory transitions transversely to the guard (i.e.\ grazing impact) or near the intersection of two guard surfaces (i.e.\ simultaneous touchdown of feet).}


In some cases, the saltation matrix for a hybrid transition can become an identity transformation.
Knowing when the saltation matrix is identity is useful to simplify computation and analysis.
The most common reason for a saltation matrix to become identity is if both of these conditions are true:
\begin{enumerate}
    \item The reset map is an identity transformation  \revise{in the neighborhood of the center of approximation, $R(x) = \id x$}, where $n$ is the dimension of the state $x$ in both $D_{\mathrm{I}}$ and $D_{\mathrm{J}}$\revise{, additionally this means that $\der_xR = I$.}
    \item The dynamics in both modes are the same before and after impact, $F^-_{\mathrm{I}} = F^+_{\mathrm{J}}$.
\end{enumerate}
\revise{With these conditions, we can see that the saltation matrix becomes the identity map:}
\begin{empheq}[box=\fbox]{align}
\arraycolsep=1pt
        \left. \begin{array}{rl}
             \revise{\der_xR_{(\mathrm{I},\mathrm{J})}} &= \id \\
             F^-_{\mathrm{I}} &= F^+_{\mathrm{J}}
        \end{array}
        \right\} \implies \Xi\revise{_{(\mathrm{I},\mathrm{J})}} = \id
         \label{eq:identitysaltation}
\end{empheq}

An example of such a transition is a foot lifting off from the ground, \revise{since there is no abrupt change in forces, the dynamics are equal at the mode transition}.
If the reset map is an identity transformation,
then $\der_x R$ is also identity and $\der_t R$ is zero. Using these conditions to simplify the expression in \eqref{eq:saltationmatrix} gives
\begin{align}
\Xi_{(\mathrm{I},\mathrm{J})} = \id+\frac{\left(F^+_{\mathrm{J}}-\id  F^-_{\mathrm{I}} - 0_{n\times n}\right)   \der_x g^-}{\der_t g^- +\der_x g^-  F^-_{\mathrm{I}}} = \id
\end{align}

\revise{Lastly, in the case that the transition is triggered by time rather than state, the saltation matrix is exactly equal to the Jacobian of the reset map $\der_x R$. This is because there is no longer a variation in the time to impact, and $\der_x g^-=0_{1\times n}$, thus
\begin{align}
\Xi_{(\mathrm{I},\mathrm{J})} = \der_x R^- \! +\frac{\left(F^+_{\mathrm{J}}\!-\der_xR^- F^-_{\mathrm{I}} \! - \der_tR^-\right)   0_{1\times n}}{\der_t g^- +0_{1\times n}  F^-_{\mathrm{I}}}  = \der_x R^-
\label{eq:timetransitionsaltation}
\end{align}
}
Therefore, in this case, it is safe to use the Jacobian of the Reset map instead of the saltation matrix, but that is because they are equivalent.
\subsection{Saltation matrix derivation}
\label{sec:derivation}
\revise{In this section, the derivation of the saltation matrix is presented, showing that \eqref{eq::saltperturbation} is satisfied by \eqref{eq:saltationmatrix}. This follows the geometric derivation from \cite{leine2004dynamics} with the addition of reset maps.}
There are alternate ways to \revise{perform this derivation and} a derivation using the chain rule is included in Appendix \ref{appendix:chainrule}.
\begin{figure*}
    \centering
    \vspace{1em}

    \begin{tikzpicture}[scale=1.2,thick]
            \node[anchor=south west,inner sep=0] at (1,0) {\includegraphics[width=0.8\textwidth]{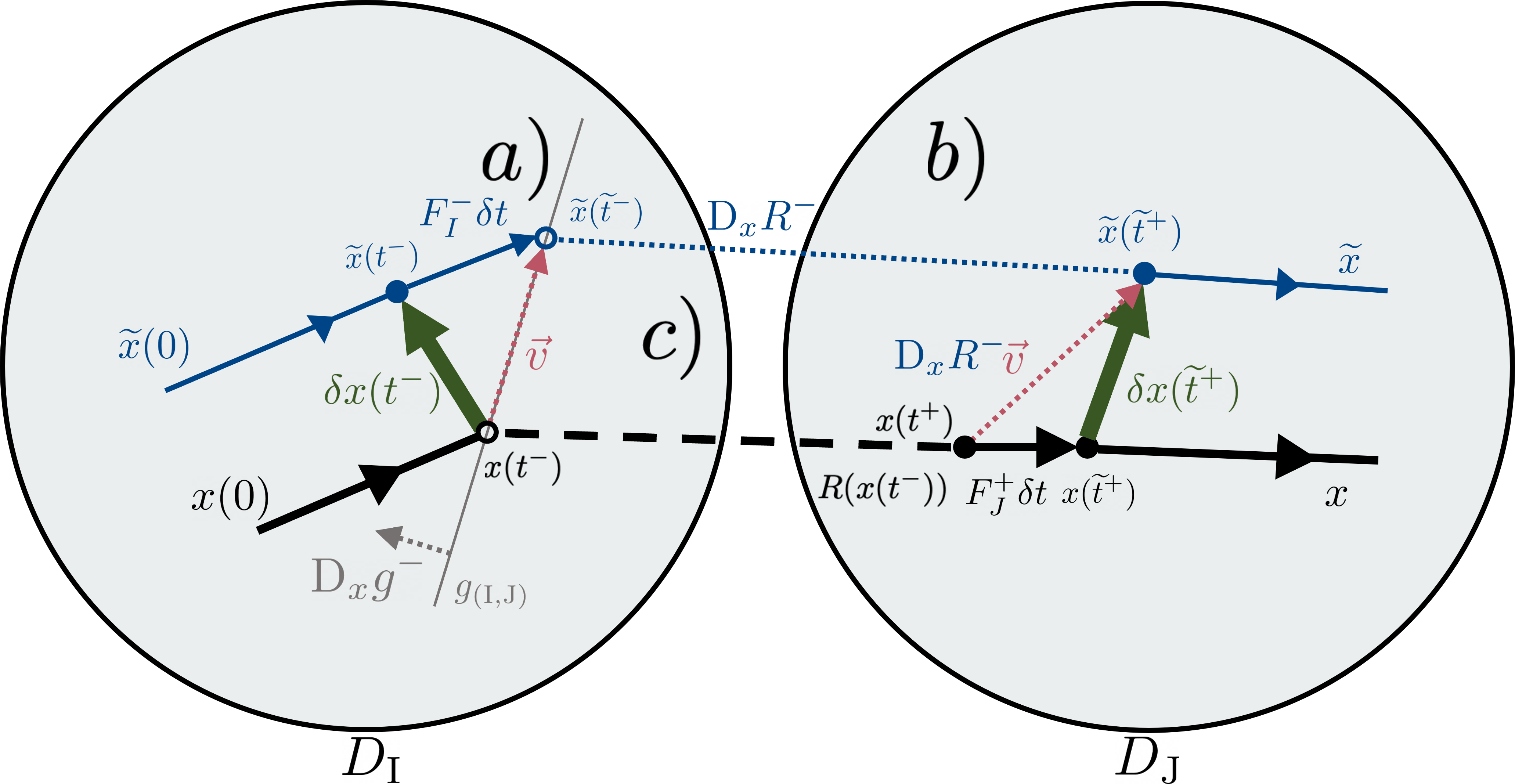}};
  \node[font = {\Large\bfseries\sffamily}] at (0.5, 0) {%
  $\begin{aligned}
     &a) &&\color{myRed}\vec{v} = \color{myBlue}F_{\mathrm{I}}^-\delta t + \color{myGreen}\delta x(t^-)\\
     &b) &&\color{myGreen} \delta x (\tilde{t}^+) = \color{myBlue}\der_x R^-\color{myRed}\vec{v} \color{black}- F_{\mathrm{J}}^+ \delta t\\
     &c) &&0= \color{myGrey}\der_xg^- \color{myRed}\vec{v}\\
  \end{aligned}$};
    \end{tikzpicture}
        \caption[Linearized hybrid system]{Linearizations made about the nominal trajectory shown in black where a perturbation is shown in \revise{green} and the perturbed trajectory is shown in blue.
    At $a)$ describes $\vec{v} = F_{\mathrm{I}}^- \delta t + \delta x(t^-)$. At $b)$ $\delta x (\pert{t}^+)$ is $\der_x R^- \vec{v} - F_{\mathrm{J}}^+ \delta t$. At $c)$  the guard condition is $0 = \der_x g^-(\delta x (t^-) + F_{\mathrm{I}}^- \delta t)$. Here $\delta t$ is positive (late transition) and for the purposes of this figure it is assumed that the system is autonomous, so the $\der_t g$ and $\der_tR$ terms drop out.
    }
    \label{fig:geometric-saltation}
\end{figure*}

Suppose the nominal trajectory of interest is $x(t)$ as shown in Fig. \ref{fig:geometric-saltation}. 
The trajectory starts in mode I and goes through a hybrid transition to mode J at time $t$.
The saltation matrix is a first-order approximation, so \revise{the dynamics are integrated with a forward Euler method. This treats the flow as a constant in each mode}, evaluated at time $t^\pm$ as in \eqref{eq:F_I_minus} and \eqref{eq:F_J_plus} such that for an infinitesimal timestep $\delta t$\revise{,}
\begin{align}
    &x(t^-) \ \revise{\approx} \ x(t^- - \delta t) + F_{\mathrm{I}}^- \delta t \qquad \textrm{in mode I} \label{eq:constant_flow}\\
    &x(t^+ +\delta t) \ \revise{\approx} \ x(t^+) + F_{\mathrm{J}}^+ \delta t\qquad \textrm{in mode J} \label{eq:constant_flow_J}
\end{align}
The reset and guard are also linearized at $t^-$ as in \eqref{eq:R_IJ_minus} and \eqref{eq:g_IJ_minus}, such that
\begin{dmath}
    \bar{R}(t^- +\delta t,x+\revise{\delta x(t^-)}) = R_{(\mathrm{I},\mathrm{J})}(t^-,x(t^-)) + \der_x R^- \revise{\delta x(t^-)}+ \der_t R^- \delta t\label{eq:reset_linearization}
\end{dmath}
\begin{dmath}
        \bar{g}(t^- +\delta t,x+\revise{\delta x(t^-)}) = g_{(\mathrm{I},\mathrm{J})}(t^-,x(t^-)) + \der_x g^- \revise{\delta x(t^-)} + \der_t g^- \delta t\label{eq:guard_linearization}
\end{dmath}
where $\bar{R}$ and $\bar{g}$ are the linear\revise{ization of the reset map and guard function about a nominal trajectory}.

Trajectories that are perturbed $\delta x$ away are labeled as $\pert{x}$.
%
\revise{Variations} can lead to changes in the impact time, which we describe with the infinitesimal time difference $\delta t := \pert{t} - t$, where $t$ is the original impact time and $\pert{t}$ is the perturbed impact time. 
If $\delta t > 0$ then the perturbed transition \revise{occurs after the nominal solution} and the \revise{perturbed} solution stays in the previous hybrid mode longer, while if $\delta t < 0$ then the perturbed solution transitions early. 
For simplicity of notation, assume the perturbed trajectory reaches the guard surface late, but the analysis also works for early transitions, resulting in the same expression \eqref{eq:saltationmatrix}, as shown in Appendix \ref{appendix:salt_derivation_early}.

Define the perturbation at the pre-impact time of the nominal trajectory $t^-$ and the post-impact time of the perturbed trajectory $\pert{t}^+$ \revise{, which we will solve for based on the initial state perturbation,} as 
%
\begin{align}
    \delta x(t^-) := \pert{x}(t^-) - x(t^-)\label{eq:delta_pert_x_t}\\
    \delta x(\pert{t}^+) := \pert{x}(\pert{t}^+) - x(\pert{t}^+)
    \label{eq:delta_x_pert_t}
\end{align}
where $\pert{x}(t^-)$ is the perturbed trajectory following the previous mode dynamics until time $t^-$.
Next, we can write \eqref{eq:delta_x_pert_t} in terms of the nominal trajectory at time of impact $x(t^-)$ and just after impact $x(t^+)$.
Using \eqref{eq:delta_pert_x_t} and \eqref{eq:constant_flow}, $\pert{x}(\pert{t}^-)$ can be written in terms of the flow before impact $F_{\mathrm{I}}^- \delta t$ and the perturbation before impact $\delta x(t^-)$:
\begin{align}
    \pert{x}(\pert{t}^-) \ =  \ x(t^-) + \delta x(t^-) + F_{\mathrm{I}}^- \delta t \revise{ + \hot} \label{eq:pert_x_pert_t}
\end{align}
Note that we denote the expression $\delta x(t^-) + F_{\mathrm{I}}^- \delta t$ as $\vec{v}$ in Fig. \ref{fig:geometric-saltation} Eq.~a.
\revise{For brevity, we will drop the higher order terms in the rest of this section.}

By using the linearized reset map \eqref{eq:reset_linearization} and the perturbation expressed in terms of the nominal trajectory \eqref{eq:pert_x_pert_t}, the reset at $\pert{x}(\pert{t}^-)$ can be \revise{expressed} in terms of the nominal state $x(t^-)$, the \revise{pre-transition} perturbation $\delta x(t^-)$, and the difference in impact time $\delta t$\revise{:}
\begin{dmath}
    \pert{x}(\pert{t}^+)
    = R(t^-,{x}(t^-)) + \der_xR^-\left(\delta x(t^-)+ F^-_{\mathrm{I}}\delta   t\right) + \der_tR^-\delta t \label{eq:barxtbar}
\end{dmath}

The final term in \eqref{eq:delta_x_pert_t} is obtained by using the constant flow after the reset \eqref{eq:constant_flow} to calculate ${x}(\pert{t}^+)$:
\begin{align}
    {x}(\pert{t}^+) &= R(t^-, x(t^-)) + F^+_{\mathrm{J}}\delta t \label{eq:x_pert_t_plus}
    \end{align}
By combining \eqref{eq:delta_x_pert_t}, \eqref{eq:barxtbar}, and \eqref{eq:x_pert_t_plus}, $\delta x(\pert{t}^+)$ can now be written as a linear function of $\delta x(t^-)$ and $\delta t$ :
\begin{align}
    \delta x(\pert{t}^+) 
    = \der_xR^- \delta x(t^-) + \left(\der_xR^- F^-_{\mathrm{I}} + \der_tR^-  -  F^+_{\mathrm{J}}\right)\delta t \label{eq:dxtpp}
\end{align}
This step is highlighted by the vector addition in Fig. \ref{fig:geometric-saltation} Eq. c.

Next, we solve for $\delta t$ as a function of $\delta x(t^-)$. 
The linear\revise{ization} of the guard \eqref{eq:guard_linearization} and the perturbation expressed in terms of the nominal trajectory \eqref{eq:pert_x_pert_t} are used to rewrite the guard evaluated at $\pert{x}(\pert{t}^-)$ as a function of the nominal \revise{solution (and noting that $ g(t^-,{x}({t}^-))  = 0$)}:
\begin{align}
    0 
    &= g(t^-,{x}({t}^-)) + \der_x g^-(\delta x(t^-)+ F_{\mathrm{I}}^- \delta t) + \der_t g^-\delta t \label{eq:gapprox}\\
     &= \der_x g^-\delta x(t^-) +(\der_x g^- F^-_{\mathrm{I}} +\der_t g^-)\delta t \label{eq:guardexpansion}
\end{align}
This expansion shows up in Fig. \ref{fig:geometric-saltation} as Eq. b.
\revise{Using \eqref{eq:guard_linearization}  to write} $\delta t$ as a function of $\delta x(t^-)$ gives
\begin{align}
    \delta t = -\frac{\der_x g^- }{\der_x g^- F^-_{\mathrm{I}} +\der_t g^-}\delta x(t^-)
    \label{eq:deltat}
\end{align}
Substituting this $\delta t$ into \eqref{eq:dxtpp} and solving for $\delta x(\pert{t}^+)$ in terms of $\delta x(t^-)$ \revise{gives}
\begin{align}
    \delta x(\pert{t}^+) &= \der_xR^-\delta x(t^-) \\
    & \quad +\frac{\left(F^+_{\mathrm{J}} - \der_xR^- F^-_{\mathrm{I}} - \der_tR^-\right)\der_x g^- }{\der_x g^- F^-_{\mathrm{I}} +\der_t g^-}\delta x(t^-)\nonumber\\
   & = \Xi_{(\mathrm{I},\mathrm{J})} \delta x(t^-) \label{eq:saltsensitivity}
\end{align}
where $\Xi$ is the saltation matrix, as in \eqref{eq::saltperturbation}.

\subsection{\revise{Gradient information using the saltation matrix}}
\label{subsec:linearforms}
\begin{figure}
    \centering
    \includegraphics[width=\linewidth]{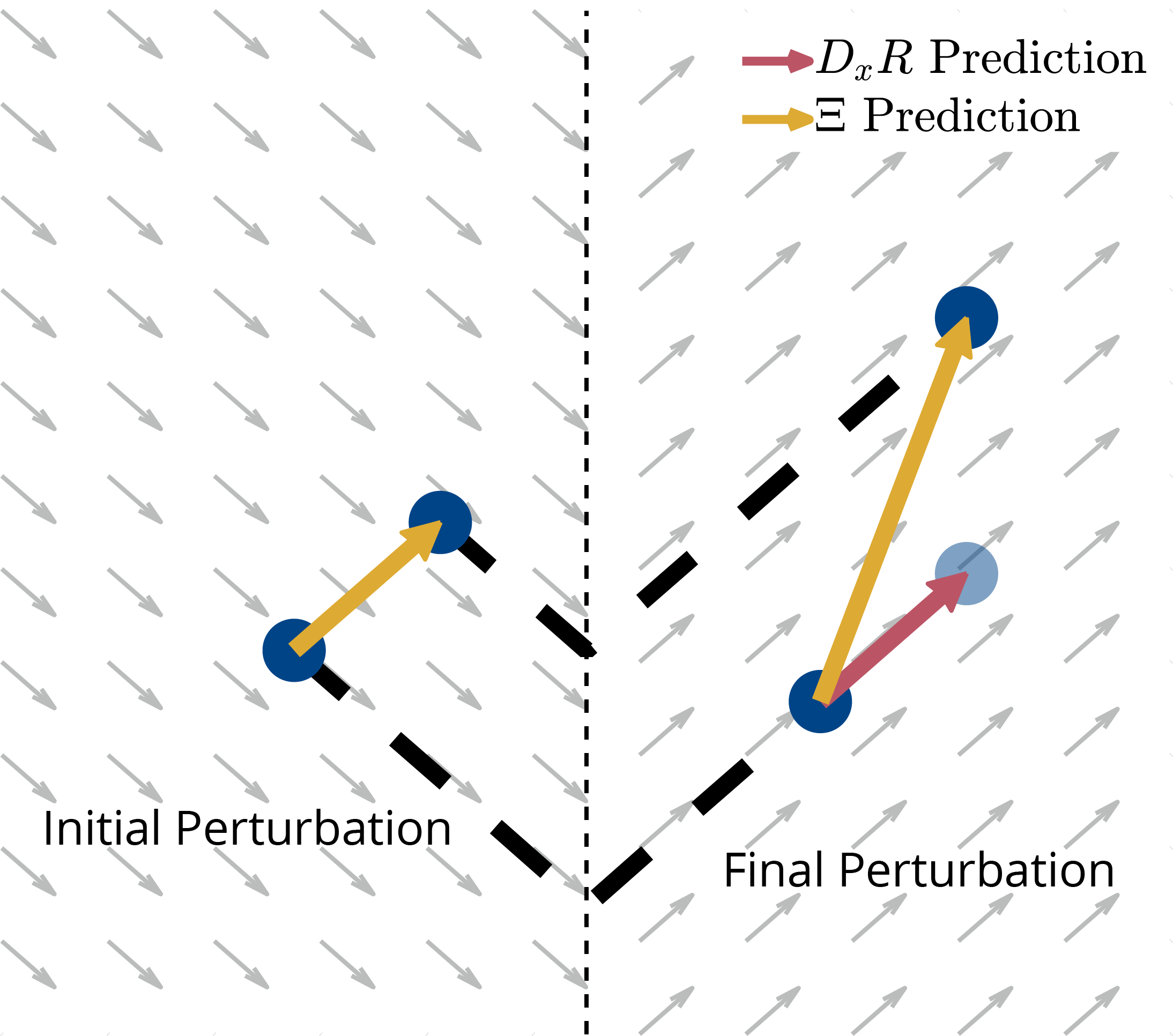}
    \caption[Flowing 2 particles through a constant flow hybrid system]{Constant flow hybrid system with identity reset map. The Jacobian of the reset map $\der_xR$ predicts no variational changes whereas using the saltation matrix $\Xi$ predicts the correct variational changes.
    }
    \label{fig:salt-gradient}
\end{figure}
Understanding how perturbed trajectories behave near a trajectory of interest is crucial for many algorithms which rely on linearizations.
The sensitivity equation describes how these \revise{variations} evolve over time. 
For a hybrid system, the time evolution simply applies the standard smooth sensitivity equation based on the Jacobian of the \revise{the dynamics with respect to state, $A_{\mathrm{I}}(t,x) := \der_x F_{\mathrm{I}}(t,x)$} \eqref{eq:sensitivity}, and the saltation matrix equation when a hybrid transition occurs \eqref{eq::saltperturbation}. 
For a transition from mode I to mode J at time $t^-$, \revise{the perturbation dynamics are described by}
\begin{align}
    \revise{\frac{d}{dt}\delta x(t)} &= A_{\mathrm{I}} \delta x (t) \:\:&s.t.&\:\: t\leq t^- \label{eq:deltaxdotminus}\\
    \delta x(t^+) &= \Xi_{(\mathrm{I},\mathrm{J})} \delta x(t^-) \:\:&s.t.&\:\: t = t^-\\
    \revise{\frac{d}{dt}\delta x(t)} &= A_{\mathrm{J}} \delta x (t) \:\:&s.t.&\:\: t \geq t^+
\end{align}
%
An example is shown in Fig.~\ref{fig:salt-gradient}, where the sensitivity is updated only by the saltation matrix because the flows are constant in both modes ($A$ is \revise{zero}).
Instead, it is the difference in mode timing that determines the change in sensitivity from the initial to final state.
If the Jacobian of the reset (which \revise{in this case is identity}) is used instead of the saltation matrix, the prediction is incorrect.
Sensitivity of hybrid systems is extensively analyzed in \cite{hiskens2000trajectory} and \cite{saccon2014sensitivity}.

Many algorithms consider finite, discrete timesteps. 
This makes the analysis slightly different, since the hybrid transition will most likely not occur exactly at the boundary of a discrete timestep.
In this case, a ``sandwich'' method is utilized, where \revise{three} (or more) smaller discrete updates are applied during a timestep in which a hybrid transition \revise{occurs}.
Consider a time interval from $t_k$ to $t_{k+1}:=t_k+\Delta$ over which a single reset occurs at time $t_k+\Delta_1$. 
The system spends $\Delta_1$ time in the first mode and $\Delta_2 := \Delta-\Delta_1$ in the second mode. 
\revise{In practice, $\Delta$ may be chosen based on a desired control update rate, while $\Delta_1$ can be solved for with a zero-crossing algorithm in an event-driven hybrid simulator or similar method.}
Let $A_{\mathrm{I},\Delta}$ be the Jacobian of the dynamics \revise{$A_{\mathrm{I}}$} discretized to time duration $\Delta$. 
Then a discrete approximation of the forward dynamics is
%
\begin{empheq}[box=\fbox]{align}
\delta x(t_{k+1}) = A_{\mathrm{J},\Delta_2} \Xi_{(\mathrm{I},\mathrm{J})} A_{\mathrm{I},\Delta_1} \delta x(t_k) \label{eq:sandwich}
\end{empheq}
which holds to first order. 
This result comes from the fundamental matrix solution \cite[Eq.~7.22]{leine2004dynamics}.
\revise{Note for the example in Fig.~\ref{fig:salt-gradient}, the constant flow in each mode means that $A_{\mathrm{I},\Delta_1} = A_{\mathrm{J},\Delta_2} = I$.}
\revise{If multiple (but finitely many) hybrid transitions occur over a time interval, additional $A_{\Delta}$ and $\Xi$ terms can be appended to \eqref{eq:sandwich} as necessary.}

Extending this idea, consider a periodic orbit of period $T$, such that $x(t)=x(t+T)$. In this case,  the fundamental matrix solution is called the monodromy matrix. If the orbit passes through modes labeled $i=\revise{\mathrm{I},\mathrm{J},\mathrm{K},...,\mathrm{Z}}$, with mode periods $T_i$ \revise{  such that $ T=\textstyle \sum_i T_i$}, then we define the \textbf{monodromy matrix} $\Phi$, \cite[Eq.~7.28]{leine2004dynamics}, \cite[Eq.~1]{wang2001monodromy}, and \cite[Eq.~12]{asahara2018stability} \revise{as}
\begin{align}
     &\revise{\Phi := \Xi_{(\mathrm{Z},\mathrm{I})}A_{\mathrm{Z},T_{\mathrm{Z}}}\cdots\,\Xi_{(\mathrm{J},\mathrm{K})}A_{\mathrm{J},T_{\mathrm{J}}} \Xi_{(\mathrm{I},\mathrm{J})}A_{\mathrm{I},T_{\mathrm{I}}}}  \label{eq:monodromydef}
    \\
     &\delta x(t+T) = \Phi \delta x(t)
     \label{eq:monodromy}
\end{align}
which holds to first order. 
%
This monodromy matrix captures the change in \revise{variations} from one cycle through the orbit to the next and \revise{its} eigenvalues (called \textbf{Floquet multipliers} \cite{leine2004dynamics}) determine the stability of the trajectory. 
Namely, if the eigenvalues all have magnitude less than one then the \revise{reference point is asymptotically stable ($\delta x(t)$ is driven to zero)}  \cite{leine2004dynamics}.  

Related to the monodromy matrix, a common technique to analyze stability of periodic systems is to analyze the return/Poincaré map \cite{leine2004dynamics}.
A \textbf{Poincaré map} $\revise{\mathcal{P}}$ converts the continuous-time system to a discrete map.
For an autonomous system with $n$ states and a limit cycle $L$, the Poincaré map is defined about a point $x^*$ on $L$ and an $n-1$ dimensional hyper-plane transverse to the flow $F$ called the Poincaré section $S$, with $x^*\in S$ \revise{and $x^*$ a fixed point of $\mathcal{P}$}.
The Poincaré map captures how points move along the Poincaré section after one cycle (\revise{$\mathcal{P}:S \to S$}). 
Stability of the fixed point is often computed by taking the Jacobian of the Poincaré map and analyzing its eigenvalues.
If \revise{all eigenvalues are within the unit circle} (the requirements for stability for a discrete system), the fixed point $x^*$ is stable.
\revise{Note that the Poincaré map only considers the state in which a trajectory crosses the Poincaré section and is not generally a function of time.
Asymptotic stability of the fixed point indicates perturbed trajectories will converge to the limit cycle, but a constant phase offset may persist ($\delta x(t)$ can never be driven to zero, only to some finite limit). As such, variations along the direction of flow along the limit cycle are invariant for autonomous systems.}

For the autonomous case, the dimension of the system is reduced by one due to the embedding \revise{of the $n-1$ dimensional Poincaré section into the state space of dimension $n$.}
\revise{On the other hand, non-autonomous systems depend explicitly on time and the Poincaré map must be augmented to consider this time dependency \cite[Ch. 9.1]{leine2004dynamics}.}
\revise{To do this}, the trajectory is augmented with a periodic time coordinate on $S^1$, and the Poincaré section is now defined to be at the end of each period $T$.
In this case, the Poincaré map and its Jacobian are in the full $n$ space, as the Poincaré section is defined on the added time coordinate.

Consider a monodromy matrix for a cycle that starts and ends at the fixed point $x^*$ for one cycle.
In the autonomous case, the monodromy matrix has the same eigenvalues as the Jacobian of the Poincaré map with an additional eigenvalue equal to \revise{1}.
This is because the monodromy matrix is still in the full $n$ space and, \revise{as mentioned above,} \revise{variations} along the direction of the flow are invariant.
\revise{For autonomous systems, the limit cycle is asymptotically stable if all other eigenvalues are within the unit circle.}
In the non-autonomous case, the monodromy matrix and the Jacobian of the Poincaré map are equivalent,
so sometimes the monodromy matrix is defined simply to be the Jacobian of the Poincaré map \cite{aroudi2015review}.

If the system is autonomous and periodic, using the Poincaré map might be more practical because the analysis is simplified by the reduction of a state variable, e.g.\ as shown for passive dynamic walkers \cite{mcgeer1990passive}.
However, the monodromy matrix \revise{is a more natural choice for non-autonomous systems and can express a stronger level of asymptotic stability where all variations from the reference point are driven to zero. The monodromy matrix can also} be generalized to the fundamental matrix solution for analysis of non-cyclical behaviors \cite[Ch. 7]{leine2004dynamics}, which the Poincaré map \revise{cannot}.
This is especially important when designing dynamic behaviors \revise{like parkour or dynamic grasps where transient growth of perturbations may cause systems to fail prior to asymptotic convergence}.

%

Also closely related to Floquet multipliers are Lyapunov exponents \cite{bockman1991lyapunov,kunze2000lyapunov,ageno2005lyapunov}. \revise{A given Floquet multiplier $\sigma$ can be written in the form} $\sigma = e^{\mu T}$ where $\mu$ is the Floquet exponent and the real part of $\mu$ is the Lyapunov exponent \cite{chicone2006ode}. If all Lyapunov exponents are negative, \revise{then} $\sigma < 1$ and the trajectory is asymptotically stable.

\subsection{\revise{Propagation of covariances and value approximations with the saltation matrix}}
\label{sec:quadraticforms}
Similar to \revise{the linear gradient forms from the last section}, quadratic forms are often used in algorithms which rely on linearizations.
Examples of such algorithms include the well-known Kalman filter and LQR controller\revise{, where quadratic forms are used to propagate the covariance distribution and value function approximation, respectively.
More formally, these equate to the propagation of the quadratic forms of vectors and co-vectors, respectively, \cite[Ch. 3]{mathai_quadratic}.
This propagation allows for accurate updates to state estimates and control laws along a trajectory.
In this context, the vector in question is the state vector, with the corresponding quadratic form being the covariance distribution. The co-vector, which lies in the space of linear functions of the vector, does not have an explicit representation here, but the quadratic form of the co-vector is the value function approximation of LQR \cite{rijnen2015optimal}.}

For covariances, consider the state trajectory as a random variable $X(t)$ with mean $\rho(t)=x(t)$, the nominal trajectory $x(t)$, and covariance $\Sigma(t)$. Define a perturbation as a zero mean random variable $\delta x(t)$ with the same covariance, such that $X(t) = x(t) + \delta x(t)$, \revise{where both $X(t)$, $x(t)$, and $\delta x(t)$ evolve according to the dynamics of the hybrid system. Therefore, once $X(0)$ and $\delta x(0)$ are sampled, the dynamics evolve deterministically.} 

\begin{figure}
    \centering
    \includegraphics[width=\linewidth]{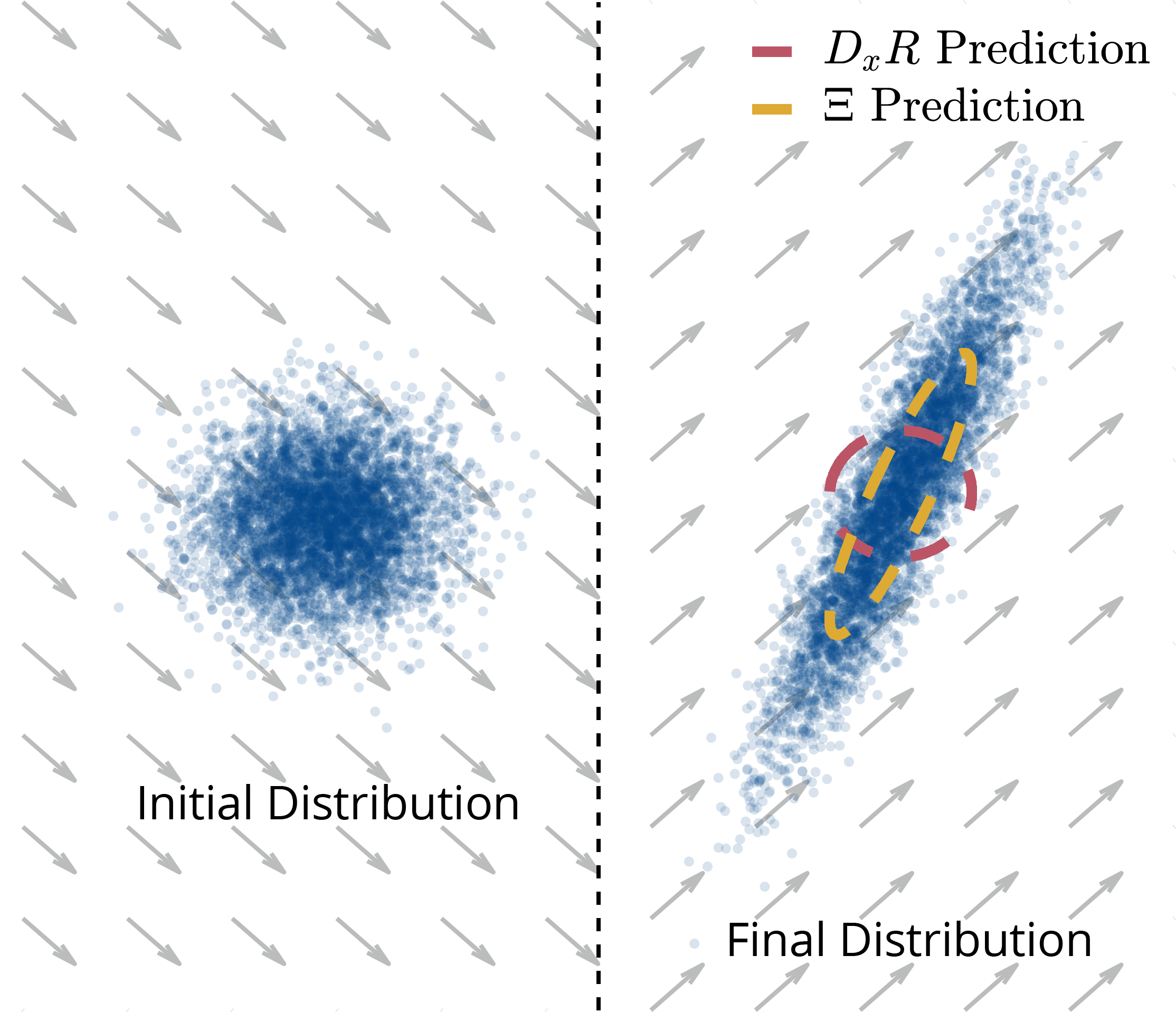}
    \caption[Flowing a distribution of particles through a constant flow hybrid system]{Constant flow hybrid system with identity reset map. The Jacobian of the reset map $\der_xR$ predicts no covariance change whereas using the saltation matrix $\Xi$ predicts the correct covariance.
    }
    \label{fig:salt-covariance}
\end{figure}

Recall that the update law for covariance \revise{$\Sigma_{\mathrm{I}}$ through a domain $\mathrm{I}$, discretized with timesteps $\Delta$, is}
\begin{align}
    \revise{\Sigma_{\mathrm{I}}(t_{k+1})} &= \revise{A_{\mathrm{I},\Delta} \Sigma_{\mathrm{I}}(t_{k}) A^T_{\mathrm{I},\Delta}}
    \label{eq:covariance_prop}
\end{align}
e.g.\ as in \cite[Eqn.~1.10]{welch1995introduction} or \cite[Eqn.~6]{julier2004unscented}, \revise{where $A_{\mathrm{I},\Delta}$ is as defined in the previous subsection}.
Similarly, at hybrid transitions, the saltation matrix applies in an analogous way (see derivation in Appendix \ref{appendix:covariance}):
\begin{empheq}[box=\fbox]{align}
    \Sigma(t^+) &= \Xi_{(\mathrm{I},\mathrm{J})} \Sigma(t^-) \Xi_{(\mathrm{I},\mathrm{J})}^T \label{eq:covarianceupdate}
\end{empheq}
\cite[Eqn.~17]{biggio2014accurate}, \cite[Eqn.~7]{paper:kong-skf-2021},
which holds to first order.
As with linear forms, the sandwich method \eqref{eq:sandwich} can be applied to retrieve the covariance propagation for an entire discrete timestep:
\begin{empheq}[box=\fbox]{align}
\Sigma(t_{k+1}) = A_{\mathrm{J},\Delta_2}\Xi_{(\mathrm{I},\mathrm{J})} A_{\mathrm{I},\Delta_1}\Sigma(t_k) A_{\mathrm{I},\Delta_1}^T\Xi_{(\mathrm{I},\mathrm{J})}^TA_{\mathrm{J},\Delta_2}^T 
\end{empheq} 
\cite[Eqn.~19]{paper:kong-skf-2021}.
An example is shown in Fig. \ref{fig:salt-covariance}, where the covariance is once again updated only by the saltation matrix because the flows are constant in both modes (\revise{$A_{\Delta}$ terms are identity}).
If the Jacobian of the reset is used instead, the incorrect covariance is predicted.
Algorithms, such as a Kalman filter \cite{paper:kong-skf-2021}, that propagate covariances with the dynamics can utilize this update law.

In the case of propagating a quadratic form of a co-vector\revise{, where the co-vector $p$ relates to the co-vector quadratic form $p = P\delta x$}, the matrix transpose terms flip sides similar to how a co-vector quadratic form propagates in the smooth domain:
\begin{align}
    P(t_{k}) = A^T_\Delta P(t_{k+1}) A_\Delta
\end{align}
as in \cite[Eqn.~3.40]{bertsekas2012dynamic}.
\revise{This structure compared with \eqref{eq:covariance_prop} highlights the dual nature of the co-vectors and vectors.}
The co-vector propagation law for the hybrid transition uses the saltation matrix in an analogous way (see derivation in Appendix \ref{appendix:Riccati}):
\begin{empheq}[box=\fbox]{align}
    P(t^-) = \Xi_{(\mathrm{I},\mathrm{J})}^T P(t^+) \Xi_{(\mathrm{I},\mathrm{J})} \label{eq::saltriccatiupdate}
\end{empheq}
\cite[Eqn.~31]{paper:kong-ilqr-2021},\cite[Eqn.~23]{rijnen2015optimal}, \revise{which is the co-vector dual to \eqref{eq:covarianceupdate}}.
The main application of the co-vector case is in the update to the Riccati equation or Bellman \revise{value function} update, e.g.\ in LQR \cite{paper:kong-ilqr-2021,rijnen2015optimal}.

\section{Example: Calculating the saltation matrix for a ball dropping on a slanted surface}
\label{sec:salt_example}
One of the simplest \revise{toy} examples of a hybrid system is a 2D point mass (ball) falling and hitting a flat surface, as shown in Fig. \ref{fig:ball-drop-covariance}.
Intuitively, the impact should eliminate variations normal to the constraint in both position and velocity.
This section presents the computation of the saltation matrix for \revise{two versions of this example: frictionless sliding and fully constrained motion due to static friction, and how both cases confirm the collapse of variations normal to the constraint.}
\revise{The insights from the section will aid in understanding the calculation of the saltation matrix for a more general class of rigid body systems described in the following section.}

\subsection{Dynamics definition}
\label{sec:example_dynamics}
Here, the system's dynamics are summarized for the 2D point mass, with an in-depth derivation for a general rigid body system given in Sec.~\ref{sec:saltcommon}.
The horizontal and vertical positions as well as their velocities are defined to be the states of the system, $x = [q^T,\dot{q}^T]^T = [q_1,q_2,\dot{q}_1,\dot{q}_2]^T$.
The ball has mass $m$ and \revise{ the magnitude of the acceleration due to gravity is $a_g$}. 
For the sake of demonstrating how inputs are handled, the ball is fully actuated with control inputs along the configuration coordinates $[u_1,u_2]^T$. \revise{In this section, we consider only plastic impact, though in Sec.~\ref{sec:elastic} we consider elastic impact.}
Two cases of friction are considered, one that assume frictionless sliding when in contact with the surface (i.e.\ the kinetic friction coefficient is zero, $\mu_k=0$) and one where the friction is sufficient to prevent sliding, i.e.\ the ball sticks to a spot. 
\revise{For these cases we define three domains: the unconstrained mode U, the constrained sliding mode S (where the ball can slide tangentially along the constraint surface), and a fully constrained mode C (where velocity is zero).}

\revise{Starting with the first case}, the ball impacts a sloped surface parameterized by an angle $\theta$, where the position constraint is defined by the guard function
\begin{align}
    g_{(\mathrm{U},\mathrm{S})}(t,x) := \sin{(\theta)} q_1 + \cos{(\theta)} q_2 =0 \label{eq:gus}
\end{align}
The resulting velocity constraint Jacobian $J_{\mathrm{S}}$ in the sliding mode is
\begin{align}
    J_{\mathrm{S}}(q) := \der_q g_{(\mathrm{U},\mathrm{S})}(t,x) = \begin{bmatrix}
    \sin{(\theta)} & \cos{(\theta)}
    \end{bmatrix} \label{eq::velcons}
\end{align}
\revise{which enforces the constraint $J_{\mathrm{S}} \dot{q} = 0$ so that the velocity must be along the surface}.
The unconstrained mode dynamics are defined by ballistic motion:
\begin{align}
    F_{\mathrm{U}}(t,x) := \begin{bmatrix}\dot{q}_1 & \dot{q}_2 & \frac{u_{1}}{m} & \frac{u_{2}-a_gm}{m} \end{bmatrix}^T
    \label{eq:F_I}
\end{align}

The hybrid guard for impact is defined by the constraint $g_{(\mathrm{U},\mathrm{S})}(q) \leq 0$, i.e when the constraint is met the impact occurs.
The reset map is defined by plastic impact\revise{ \cite{pfeiffer2008multibody}}, which enforces the velocity constraint:
\begin{align}
    R_{(\mathrm{U},\mathrm{S})}(t,x) := \begin{bmatrix} q_{1}\\ 
    q_{2}\\ 
    \dot{q}_1\,{\cos^2\left(\theta \right)}-\dot{q}_2\,\cos\left(\theta \right)\,\sin\left(\theta \right)\\ 
    \revise{\dot{q}_2\,{\sin^2\left(\theta \right)} -\dot{q}_1\,\cos\left(\theta \right)\,\sin\left(\theta \right)} \end{bmatrix} \label{eq:R_IJ}
\end{align}

The constrained mode dynamics are found by solving the ballistic dynamics while maintaining the velocity constraint: 
\begin{align}
    F_{\mathrm{S}}&(t,x):=  \\&\begin{bmatrix} \dot{q}_1\\ \dot{q}_2\\ \frac{u_{1}\,{\cos^2\left(\theta \right)}}{m}-\frac{u_{2}\,\cos\left(\theta \right)\,\sin\left(\theta \right)}{m}+\frac{\revise{a_g}\,m\,\cos\left(\theta \right)\,\sin\left(\theta \right)}{m}\\ 
    -\frac{u_{1}\,\cos\left(\theta \right)\,\sin\left(\theta \right)}{m} +
    \frac{u_{2}\,{\sin^2\left(\theta \right)}}{m}-\frac{\revise{a_g}\,m\,{\sin^2\left(\theta \right)}}{m} \end{bmatrix} \nonumber\label{eq:F_J}
\end{align}

In the case of sticking friction in mode $\mathrm{C}$, \revise{the guard function is equivalent to \eqref{eq:gus} ($g_{(\mathrm{U},\mathrm{C})}(t,x) = g_{(\mathrm{U},\mathrm{S})}(t,x))$, but} there is a no slip condition added to \eqref{eq::velcons}:
\begin{align}
    J_{\mathrm{C}} := \begin{bmatrix}
    -\cos{(\theta)} & \sin{(\theta)}\\
    \sin{(\theta)} & \cos{(\theta)}
    \end{bmatrix}
\end{align}
\revise{which enforces the constraint $J_{\mathrm{C}} \dot{q} = 0$ so that velocity is zero.
The constrained dynamics are}
\begin{align}
    \dot{x} = F_{\mathrm{C}}(t,x) = \begin{bmatrix}
    \revise{0} &
    \revise{0} &
    0 &
    0
    \end{bmatrix}^T
\end{align}
The reset map eliminates all velocities:
\begin{align}
    \revise{R_{(\mathrm{U},\mathrm{C})}(t,x)} := \begin{bmatrix} q_{1}&
    q_{2} &
    0 &
    0 \end{bmatrix}^T
    \label{eq:R_UC}
\end{align}
Note that \revise{the state in} this mode is fully constrained and the ball will just stick to the surface (as $\dot{q}=0$ after impact).

\subsection{Saltation matrix calculation}
To compute the saltation matrix, the Jacobians of the guard and reset map with respect to state must be computed.
The Jacobian of the guard is simply the velocity constraint Jacobian padded with zeros for each velocity coordinate:
\begin{align}
    \der_x g_{(\mathrm{U},\mathrm{S})}(t,x) = \begin{bmatrix} J_{\mathrm{S}} & 0_{1\times2} \end{bmatrix} = \begin{bmatrix} \sin{(\theta)}&\cos{(\theta)}&0&0\end{bmatrix} \label{eq:DxG}
\end{align}
The Jacobian of the reset map is
\begin{align}
    \der_x&R_{(\mathrm{U},\mathrm{S})}(t,x) =  \label{eq:DxR}\\
    &\begin{bmatrix} 1 & 0 & 0 & 0\\ 0 & 1 & 0 & 0\\ 0 & 0 & {\cos^2\left(\theta \right)} & -\cos\left(\theta \right)\,\sin\left(\theta \right)\\ 0 & 0 & -\cos\left(\theta \right)\,\sin\left(\theta \right) & \sin^2\left(\theta \right)\end{bmatrix} \nonumber
\end{align}
The saltation matrix is then computed by substituting in each component, \eqref{eq:F_I}--\eqref{eq:DxR}, into the definition, \eqref{eq:saltationmatrix}, to get
\begin{empheq}[box=\fbox]{align}
        \Xi_{(\mathrm{U},\mathrm{S})} = \begin{bmatrix} \Omega_{(\mathrm{U},\mathrm{S})} & 0_{2\times 2}\\
        0_{2\times 2} & \Omega_{(\mathrm{U},\mathrm{S})}\end{bmatrix}
        \label{eq:salt_us}
\end{empheq}
where $\Omega_{(\mathrm{U},\mathrm{S})}$ is a block element consisting of
\begin{align}
    \Omega_{(\mathrm{U},\mathrm{S})} := \begin{bmatrix}
    \cos^2\left(\theta \right) & -\cos\left(\theta \right)\,\sin\left(\theta \right)\\
            -\cos\left(\theta \right)\,\sin\left(\theta \right) & \sin^2\left(\theta \right)
    \end{bmatrix}
    \label{eq:salt_block_us}
\end{align}

\revise{Note that the control input does not appear in the saltation matrix, indicating that if the control input is constant across the hybrid transition, it has no effect on the evolution of variations across the transition. This is not necessarily true in general, as discussed in the following section.
Also note the block diagonal structure of the saltation matrix, which has interesting implications discussed in Sec. \ref{sec:exanalysis}.}

For the sticking saltation matrix, similar calculations are made as in the sliding case:
\begin{align}
    \der_x\revise{g_{(\mathrm{U},\mathrm{C})}}(t,x) = \begin{bmatrix} \sin{(\theta)} & \cos{(\theta)} & 0 & 0\end{bmatrix}
\end{align}
Note that the guard condition is the same, which results in having the same Jacobian of the guard as the sliding case.
The Jacobian of the reset map is
\begin{align}\der_xR_{(\mathrm{U},\mathrm{C})}(t,x) = \begin{bmatrix} \revise{I_{2x2}} &\revise{ 0_{2x2}}\\ \revise{0_{2x2}} & \revise{0_{2x2}}\end{bmatrix}
\end{align}
The resulting saltation matrix becomes
\begin{empheq}[box=\fbox]{align}
    \Xi_{(\mathrm{U},\mathrm{C})} = \begin{bmatrix}\Omega_{(\mathrm{U},\mathrm{C})} & 0_{2\times 2}\\0_{2\times 2} & 0_{2\times 2} \end{bmatrix} 
    \label{eq:salt_uc}
\end{empheq}
where $\Omega_{(\mathrm{U},\mathrm{C})}$ is a block element consisting of
\begin{align}
    \Omega_{(\mathrm{U},\mathrm{C})} :=  \frac{1}{{\dot{q}_{2}\,\cos\left(\theta \right)+\dot{q}_{1}\,\sin\left(\theta \right)}}
    \begin{bmatrix} \dot{q}_{2}\,\cos\left(\theta \right) &
    -\dot{q}_{1}\,\cos\left(\theta \right)  \\     -\dot{q}_{2}\,\sin\left(\theta \right) & \dot{q}_{1}\,\sin\left(\theta \right)
    \end{bmatrix}
    \label{eq:salt_block_uc}
\end{align}
\revise{Note that \eqref{eq:salt_block_uc} and, as a result, \eqref{eq:salt_uc} will go to infinity as the term ${\dot{q}_{2}\,\cos\left(\theta \right)+\dot{q}_{1}\,\sin\left(\theta \right)}$ goes to zero. This is a consequence of a non-transverse guard crossing that violates \eqref{asm:transverse}.
Also observe that the saltation matrix's dependence on state is highly non-linear, despite the linearization of the guard and reset map.}

\subsection{Saltation matrix analysis}
\label{sec:exanalysis}
\begin{figure}
    \centering
    \includegraphics[width=\linewidth]{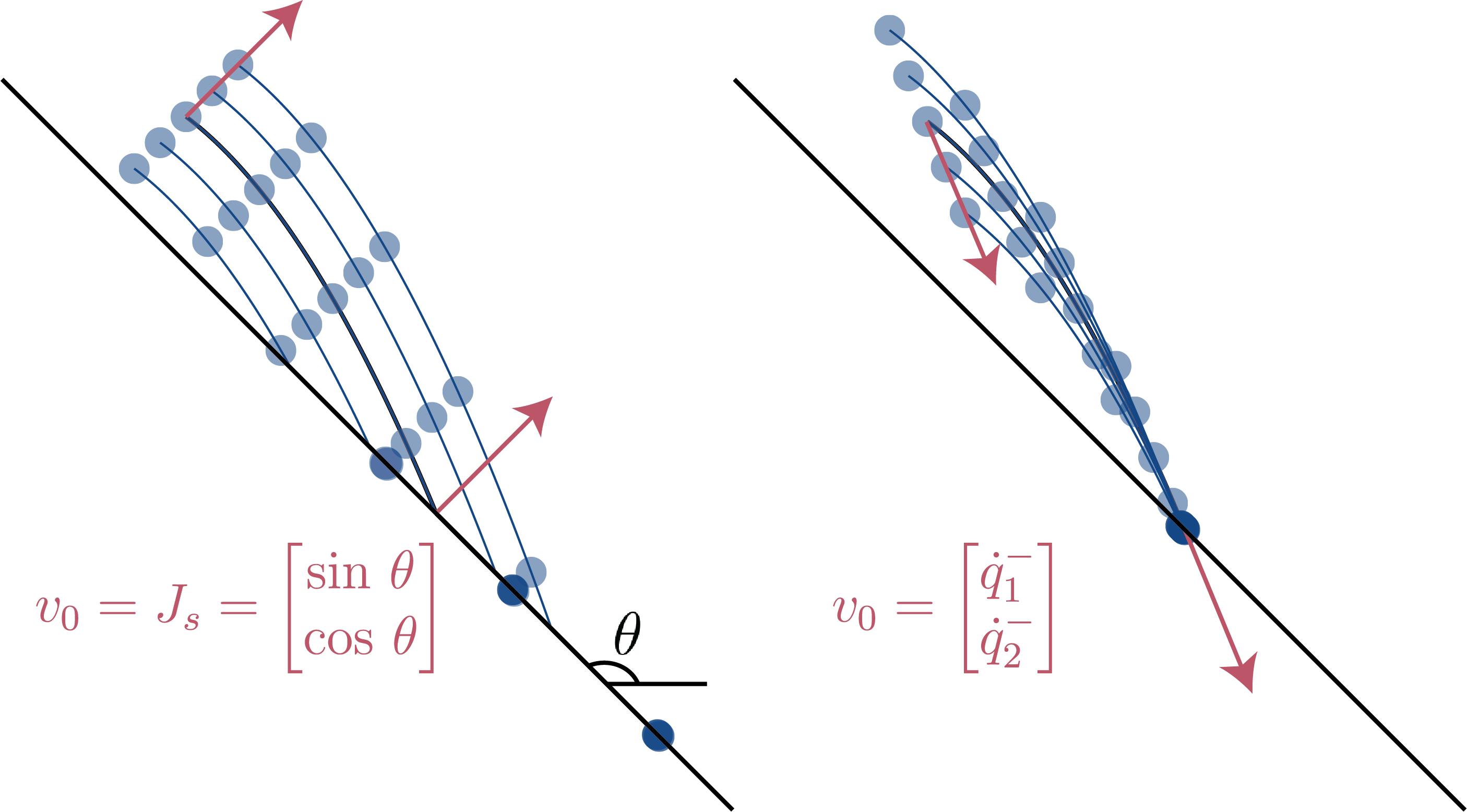}
    \caption[Eigenvector analysis of impact into sliding]{Ball drop example with sliding friction (left) and sticking friction (right). In sliding, position variations in the direction of the constraint are eliminated. \revise{The eigenvector $v_0$ is} associated with the zero eigenvalue and $\theta$ is the angle of the surface. In sticking, position variations in the direction of pre-impact velocity are eliminated. $q_1$ is the horizontal configuration, and $q_2$ is the vertical configuration.
    }
    \label{fig:ball-drop}
\end{figure}

The saltation matrix for the sliding case $\Xi_{(\mathrm{U},\mathrm{S})}$ is a block diagonal matrix with a repeating block element, shown in \eqref{eq:salt_us}--\eqref{eq:salt_block_us}.
This implies that the variations in position are mapped equivalently to variations in velocity.
The eigenvalues and corresponding eigenvectors of this block are
\begin{align}
    \lambda_0 &= 0, &\lambda_1&=1\nonumber\\
    v_0 &= \begin{bmatrix}
    \sin{(\theta)}\\
    \cos{(\theta)}
    \end{bmatrix}, &v_1&=\begin{bmatrix}
    \revise{-\cos{(\theta)}}\\
    \revise{\sin{(\theta)}}
    \end{bmatrix}
\end{align}
The first eigenvalue is zero, so any variation in the direction of its eigenvector is eliminated.
Note that this eigenvector is exactly the velocity constraint Jacobian, $J_{\mathrm{S}} = [\sin{(\theta)},\cos{(\theta)}]$.
Thus, variations along the normal direction of the constraint for both position and velocity are zeroed out, i.e.\ there are no variations normal to the surface once impact is made, as shown in Fig.~\ref{fig:ball-drop}. Note that while the reset map zeros out velocity in this direction (and so this effect arises from the $\der_xR$ term), the reset map has no effect on positions. For the position block, the effect in the constraint direction arises from the $\der_xg$ term in the numerator of the second term in \eqref{eq:saltationmatrix}, as in \eqref{eq:DxG}.

The second eigenvalue is identity, so variations in the direction of its eigenvector do not change.
This eigenvector is tangent to the constraint direction, $[-\cos{(\theta)},\sin{(\theta)}]$. 
In fact, the saltation matrix is always just a rank one update to $\der_xR$ in the direction of $\der_xg$ and all other directions are unaffected.
Although this is a simple example, this block matrix structure exists for all rigid body systems with unilateral constraints, as explored in the next section.

For plastic impact into sticking, ($\mathrm{U},\mathrm{C})$, variations in configuration map differently than velocity variations.
This is because the tangential constraint is only applied to the velocity and not the position (i.e.\ it is non-holonomic), whereas in the normal direction, both position and velocity are constrained.
The sticking saltation matrix $\Xi{(\mathrm{U},\mathrm{C})}$ reflects this change, where there is no longer a repeated element in the block diagonal.
Instead, the only nonzero component is how variations in position map onto the constraint surface  \eqref{eq:salt_uc}--\eqref{eq:salt_block_uc}.
The velocity components are all zero because velocity is fully constrained to zero.
Again, we analyze the non-zero block by computing the eigenvalues and corresponding eigenvectors:
\begin{align}
    \lambda_0 &= 0, &\lambda_1&=1\nonumber\\
    v_0 &= \begin{bmatrix}
    \dot{q}_1\\
    \dot{q}_2
    \end{bmatrix}, &v_1&=\begin{bmatrix}
    -\cos{(\theta)}\\
    \sin{(\theta)}
    \end{bmatrix}
\end{align}
Similar to the sliding case, variations tangential to the constraint are preserved. 
However, the zero eigenvector is different.
Configuration variations that are in the same direction as the impact velocity disappear.
Fig.~\ref{fig:ball-drop} illustrates this idea, where position variations in the direction of the pre-impact velocity are eliminated.
This is intuitive because the ball impacting earlier or later has no effect if the variation is in line with the impact velocity, it will hit the same contact point and stick.

\section{Saltation matrices for rigid body systems with \revise{time-invariant} unilateral constraints}
\label{sec:saltcommon}
For rigid body systems with contacts, the hybrid modes are the enumeration of different contact conditions. 
This section defines the dynamics of these systems and calculates the saltation matrix of all the common mode transitions for a single constraint.
This section \revise{expands} much of the intuition developed in Sec.~\ref{sec:salt_example} \revise{to a broader category of rigid body systems.
In particular, we identify several relevant transition cases that expose relatively straightforward structure of the corresponding saltation matrices, summarized in Table \ref{table:saltation_structure}.
There are more cases that are not addressed in this section due to additional complexity in the dynamics, that do not deliver further insight to the structure of the saltation matrix, but can be derived similarly if necessary.
Similarly, this section considers rigid body systems with time-invariant unilateral constraints, though it is straightforward to consider this time dependence, such as in the case of a paddle juggler \cite{paper:zhu-hybrid-2022}.}

\subsection{Dynamics derivation}
\label{sec:rbdynamics}
\begin{figure}
    \centering
    \includegraphics[width=0.35\textwidth]{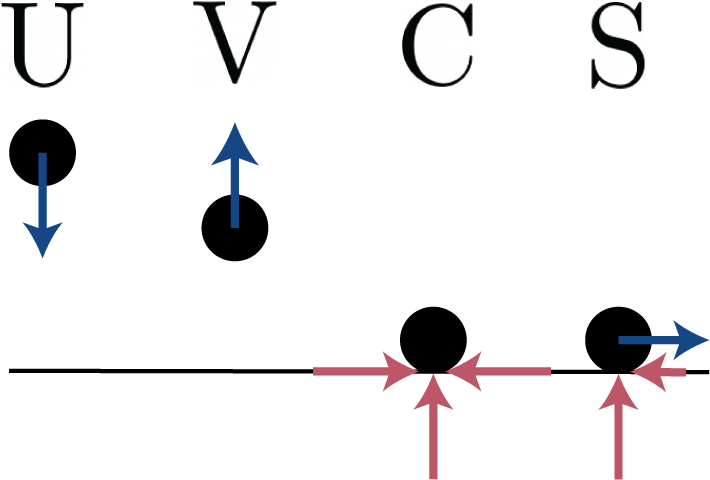}
    \caption[Rigid body hybrid modes]{Depicting the different rigid body hybrid modes considered where blue arrows depict velocities and red arrows depict forces. 
    $\mathrm{U}$ is the unconstrained mode with approaching velocity to the constraint, $\mathrm{V}$ is the unconstrained mode with separating velocity, $\mathrm{C}$ is the constrained mode, and $\mathrm{S}$ is the sliding mode on the constraint. 
    A single planar point is shown here, but the system may have additional degrees of freedom.}
    \label{fig:rigid_body_modes}
\end{figure}

The following examples consider four modes, 
illustrated in Fig. \ref{fig:rigid_body_modes}: the unconstrained mode \revise{$\mathrm{U}$ is when }approaching the constraint surface, the unconstrained mode \revise{$\mathrm{V}$ is when }leaving the \revise{constraint} surface, a constrained mode $\mathrm{C}$, and a \revise{sliding-with-friction-mode} $\mathrm{S}$. 
The reason both $\mathrm{U}$ and $\mathrm{V}$ are included is to ensure that elastic impact is not defined with a self-reset \revise{(where a self-reset is when a hybrid mode resets to itself causing a discontinuity in state within the same hybrid domain)} and to avoid degenerate impacts just after liftoff, when the velocity is not approaching the constraint but the guard condition is satisfied $g_{\mathrm{n}} \leq 0$, especially when using numerical integration.

\revise{Combinatorially, there are 12 possible transitions that could occur between these modes.
However, there are 4 degenerate transitions: $(\mathrm{V},\mathrm{C})$, $(\mathrm{V},\mathrm{S})$, $(\mathrm{C},\mathrm{U})$, and $(\mathrm{S},\mathrm{U})$, that are not realizable since the system can not impact a surface while while moving away from it nor liftoff into the surface.
The remaining 8 transitions are discussed in this section.}

The states of the system are the configuration coordinates $q$ and their velocities $\dot{q}$, such that $x := [q,\dot{q}]^T$.
The dimension of the configuration $q$ is defined to be $m$, while the dimension of the state space $x$ is $n = 2m$.
Contacts between rigid bodies are regulated through a unilateral constraint in the normal (n) direction, $g_{\mathrm{n}}(t,x)\geq 0$.
Note that $g_{\mathrm{n}}(t,x)$ only depends on the configuration $q$ and not the velocity. 
When rigid bodies are in contact they must satisfy $g_{\mathrm{n}}(t,x) = 0$.

The Jacobian of $g_{\mathrm{n}}$ with respect to the configuration coordinates is defined to be $J_{\mathrm{n}}\revise{(t,q)} := \der_q g_{\mathrm{n}}(t,x)$.
In the sliding mode, the constraint Jacobian consists of just this normal direction constraint, $J_{\mathrm{S}}=J_{\mathrm{n}}$. However, if the no slip condition is added, the constrained mode C has a constraint Jacobian of
\begin{align}
    J_{\mathrm{C}}\revise{(t,q)} := \begin{bmatrix}
    J_{\mathrm{n}}\revise{(t,q)}\\
    J_{\mathrm{t}}\revise{(t,q)}
    \end{bmatrix}
\end{align}
where $J_{\mathrm{t}}\revise{(t,q)}$ is the tangential velocity constraint Jacobian.
For unconstrained modes, \revise{$J \in \mathbb{R}^{0\times 0}$} is empty.

\revise{Define that $J(t,q)$ with no subscript specifies any mode, where it can be empty, just the normal, or both the normal and tangential component. We also assume that $J(t,q)$ is differentiable with respect to time.}
In any mode, the following acceleration constraint is applied based on $J\revise{(t,q)}$ for that mode to maintain the active constraints until the next guard
\begin{equation}
    J(t,\revise{q})\ddot{q} + \dot{J}(t,\revise{q})\dot{q} = 0 \label{eq:accelcons}
\end{equation}
\revise{This acceleration constraint is derived by differentiating the velocity constraint once with respect to time using the chain rule.}

The equations of motion for each mode are defined by the constrained manipulator dynamics, e.g.\ \cite{murray2017mathematical}, where this constraint is combined with Lagrangian dynamics:
\begin{equation}
    \begin{bmatrix}
    M&J^T\\
    J&0
    \end{bmatrix}\begin{bmatrix} 
    \ddot{q} \\ f 
    \end{bmatrix} = \begin{bmatrix}
    \Upsilon-N\\0
    \end{bmatrix}-\begin{bmatrix}
    C\\\dot{J}
    \end{bmatrix}\dot{q}
    \label{eq:blockdynamics}
\end{equation}
\cite[Eqn.~33]{johnson2013legged} where $f$ is the constraint force vector (Lagrange multiplier), $M(q)$ is the mass matrix, $C(q,\dot{q})$ is the Coriolis matrix, $\Upsilon(u)$ the input vector, and $N(q,\dot{q})$ are the other nonlinear forces such as gravity and sliding friction. 

To help with the following equations, \revise{we use }the $\dagger$ notation from \cite[Eqn.~8]{johnson2016hybrid} \revise{to label the blocks of the following matrix inverse}, where in each mode:
\begin{align}
\begin{bmatrix}
 M^\dagger& J^{\dagger T}\\
  J^{\dagger}& \Lambda^\dagger \\
\end{bmatrix} := \begin{bmatrix}
    M&J^T\\
    J&0
    \end{bmatrix}^{-1}
    \label{eq::dagger}
\end{align}
This definition produces a number of identities, in particular\revise{,}
\begin{align}
M^\dagger M = \idm - J^{\dagger T}J \label{eq:mdaggeridentity}
\end{align}
\cite[Eqn.~11]{johnson2016hybrid}, which will be helpful in simplifying the saltation matrix expressions. \revise{Note that in the unconstrained case, $M^{\dagger} = M^{-1}$, $J^{\dagger}\in \mathbb{R}^{0\times0}$, and $\Lambda^{\dagger}\in \mathbb{R}^{0\times0}$.}

With this notation, the state space dynamics can be \revise{solved by multiplying the matrix inverse to the right side of \eqref{eq:blockdynamics} and is} expressed as
\begin{align}
    \dot{x} = \frac{d}{dt} \begin{bmatrix} q \\ \dot{q} \end{bmatrix} = \begin{bmatrix}
    \dot{q}\\
    M^\dagger \left(\Upsilon-N-C\dot{q}\right) - J^{\dagger T}\dot{J}\dot{q}
    \end{bmatrix}
    \label{eq::dynamics}
\end{align}
\cite[Eqn.\ 75]{johnson2016hybrid} where each $\dagger$ component is different depending on the hybrid mode based on $J$.

\revise{By similarly multiplying the matrix inverse to the right side of \eqref{eq:blockdynamics}}, the constraint forces $f(t,x)$ are calculated from the bottom row of \eqref{eq:blockdynamics}:
\begin{align}
    f(t,x) = J^\dagger \left(\Upsilon-N-C\dot{q}\right) - \Lambda^\dagger\dot{J}\dot{q}
    \label{eq:fdynamics}
\end{align}

Coulomb friction is used in the sliding mode -- frictional forces in the tangential direction $f_{\mathrm{t}}$ (included in $N$) are applied to resist sliding motion proportional to the normal constraint force, $f_{\mathrm{n}}$, and in the direction resisting the sliding velocity, $v_{\mathrm{t}} = J_{\mathrm{t}}\dot{q}$:
\begin{align}
f_{\mathrm{t}} =\mu_k f_{\mathrm{n}} \frac{J_{\mathrm{t}}\dot{q}}{\|J_{\mathrm{t}}\dot{q}\|} = \mu_k f_{\mathrm{n}} \frac{v_{\mathrm{t}}}{\|v_{\mathrm{t}}\|} \label{eq:coulomb}
\end{align}
where $\mu_k$ is the kinetic coefficient of friction. 

When a contact constraint is added, for example the normal surface constraint $g_{\mathrm{n}}$,
an impact law $J_{\mathrm{n}}\dot{q}^+ = -eJ_{\mathrm{n}}\dot{q}^-$ is applied (where the coefficient of restitution $e=1$ is perfectly elastic and $e=0$ is perfectly plastic)
along with the impulse momentum equation to get
  \begin{equation}
    \begin{bmatrix} 
    \dot{q}^+ \\ \hat{p}
    \end{bmatrix} = \begin{bmatrix}
    M&J_{\mathrm{n}}^T\\
    J_{\mathrm{n}}&0
    \end{bmatrix}^{-1}\!\!\begin{bmatrix}
    M\\-eJ_{\mathrm{n}}
    \end{bmatrix}\dot{q}^- = \begin{bmatrix}
 M^\dagger_{\mathrm{n}}& J^{\dagger T}_{\mathrm{n}}\\
  J^{\dagger}_{\mathrm{n}}& \Lambda_{\mathrm{n}}^\dagger \\
\end{bmatrix}\!\!\begin{bmatrix}
    M\\-eJ_{\mathrm{n}}
    \end{bmatrix}\dot{q}^-
    \label{eq::resetmapdagger}
\end{equation} 
where $\hat{p}$ is the impulse magnitude vector \revise{\cite[Eqn.~23]{johnson2016hybrid}, \cite{johnson2021impact}}.
Since the positions do not change instantaneously, the state space reset map for elastic, frictionless impact from mode U to mode V is
\begin{equation}
    x^+ = \begin{bmatrix}
    q^+\\
    \dot{q}^+
    \end{bmatrix} = R_{(\mathrm{U,V})}(t,x^-) = \begin{bmatrix}
    q^-\\
    M^\dagger_{\mathrm{n}} M \dot{q}^- - eJ^{\dagger T}_{\mathrm{n}}J_{\mathrm{n}}\dot{q}^-
    \end{bmatrix}
    \label{eq::elasticresetmap}
\end{equation}
The plastic, frictionless impact reset map into mode S follows \eqref{eq::elasticresetmap} but with $e=0$ (and written with $M^{\dagger}_{\mathrm{S}}$ for mode S, though 
$M^\dagger_{\mathrm{S}} = M^\dagger_{\mathrm{n}}$ since 
$J_{\mathrm{S}}=J_{\mathrm{n}}$):
\begin{equation}
    x^+ = \begin{bmatrix}
    q^+\\
    \dot{q}^+
    \end{bmatrix} = R_{(\mathrm{U},\mathrm{S})}(t,x^-) = \begin{bmatrix}
    q^-\\
    M^\dagger_{\mathrm{S}} M \dot{q}^-
    \end{bmatrix}
    \label{eq::plasticresetmap}
\end{equation}
The frictional, plastic impact reset map, $R_{(\mathrm{U},\mathrm{C})}$, follows \eqref{eq::plasticresetmap} but with $J_{\mathrm{C}}$ and $M^\dagger_{\mathrm{C}}$ instead of $J_{\mathrm{S}}$ and $M^\dagger_{\mathrm{S}}$. 
Similarly, the liftoff reset maps into modes U or V are the same except that there is no constraint $J$, and so the reset simplifies to an identity map. 
Note that the reset map does not depend on the prior mode, so for example $R_{(\mathrm{S,C})}=R_{(\mathrm{U},\mathrm{C})}$.

\subsection{Apex}
Apex is a ``virtual'' hybrid event -- one that does not have a physical reset map or change in the dynamics -- and is triggered when the velocity switches from going away from the constraint to towards the constraint $(\mathrm{V,U})$.
As the reset map is identity, and the dynamics match before and after (since there is not a difference in control at this event) the saltation matrix is identity following \eqref{eq:identitysaltation}:
\begin{align}
    \Aboxed{\Xi_{(\mathrm{V,U})} = \id}
    \label{eq:salt_v_u}
\end{align}

\subsection{Liftoff}
Liftoff is a hybrid transition into mode $\mathrm{V}$ from $\mathrm{S}$ or $\mathrm{C}$ that depends on the constraint force $f(t,x)$, \revise{defined in }\eqref{eq:fdynamics}, which is a function of both time and state (and implicitly a function of control input).
\revise{Note that liftoff and apex cannot occur at the same time due to the transversality requirement.}
The guard for liftoff is determined by $f_{\mathrm{n}}$, the constraint force in the $J_{\mathrm{n}}$ direction -- if the force becomes non-repulsive, then the contact is released:
 \begin{align}
     g_{(\mathrm{C,V})}(t,x) := f_{\mathrm{n}}(t,x), \quad
     g_{(\mathrm{S,V})}(t,x) := f_{\mathrm{n}}(t,x)
 \end{align}
Because the hybrid event occurs when the constraint force goes to zero, the dynamics at the boundary are equal. 
This is true even in the case of sticking friction in mode $\mathrm{C}$, as the friction cone ensures that either the system transitions to sliding mode $\mathrm{S}$ (as discussed in Sec.~\ref{sec:friction}) or the frictional force goes to zero at the same time.  
The state does not jump during liftoff, which meaning the reset map for liftoff is an identity transformation.
Since both conditions of \eqref{eq:identitysaltation} are met for liftoff, the saltation matrices are identity:  
\begin{align}
    \Aboxed{\Xi_{(\mathrm{C,V})} &= \id} \label{eq:salt_c_v}\\
    \Aboxed{\Xi_{(\mathrm{S,V})} &= \id} \label{eq:salt_s_v}
\end{align}
Due to the smooth nature of liftoff, these events can be safely ignored when considering variations from liftoff.

\subsection{Plastic impact}
Plastic impact occurs when the unconstrained mode $\mathrm{U}$ makes contact and transitions to either the sliding mode $\mathrm{S}$ or the constrained mode $\mathrm{C}$.
First, consider plastic impact into sliding $(\mathrm{U},\mathrm{S})$.
For simplicity, frictionless sliding $\mu_k = 0$ is assumed to expose the structure in the saltation matrix, but the same calculations can be made with non-zero sliding friction $\mu_k > 0$.
The dynamics for each mode is from \eqref{eq::dynamics}:
\begin{align}
    F_{\mathrm{U}}(t,x^-) &= \begin{bmatrix}
    \dot{q}^-\\
    M^{-1}(\Upsilon - C^-\dot{q}^--N)
    \end{bmatrix}\\
    F_{\mathrm{S}}(t,x^+) &= \begin{bmatrix}
    \dot{q}^+\\
    M_{\mathrm{S}}^{\dagger}(\Upsilon - C^+\dot{q}^+-N) - J_{\mathrm{S}}^{\dagger T}\dot{J}_{\mathrm{S}}^+\dot{q}^+
    \end{bmatrix}
\end{align}
Note that $-$ or $+$ on $\mathrm{C}$ and $\dot{J}$ indicates that these functions use the pre- or post-impact velocity, $\dot{q}^-$ or $\dot{q}^+$, respectively\revise{: i.e. $C^- := C(t, x^-)$ and $C^+ := C(t, R(t, x^-))$}.
The Jacobian of the reset map for plastic impact, \eqref{eq::plasticresetmap}, is
\begin{align}
    \der_xR_{(\mathrm{U},\mathrm{S})}(t,x^-) = \begin{bmatrix}
    \idm & 0_{m\times m}\\
    \der_q(M_{\mathrm{S}}^\dagger M \dot{q}^-) & M_{\mathrm{S}}^\dagger M
    \end{bmatrix} 
    \label{eq:resetmapjacobianplastic}
\end{align}
The Jacobian of the guard $\der_xg_{(\mathrm{U},\mathrm{S})}(t,x^-)$  is
\begin{align}
    \der_xg_{(\mathrm{U},\mathrm{S})}(t,x^-) &= \begin{bmatrix}J_{\mathrm{S}} & 0_{1\times m}\end{bmatrix}
\end{align}
while the denominator of $\Xi_{(\mathrm{U},\mathrm{S})}$ is the impact velocity:
\begin{align}
    \der_x g_{(\mathrm{U},\mathrm{S})}(t,x^-) F_{\mathrm{U}}(t,x^-) &= \begin{bmatrix}J_{\mathrm{S}} & \!0_{1\times m}\end{bmatrix}F_{\mathrm{U}}(t,x^-) =  J_{\mathrm{S}} \dot{q}^-
\end{align}

\revise{As we are considering time invariant systems}, the guard and reset map are independent of time, $\der_t R = 0_{n\times 1}, \der_t g = 0$.
However, in other cases such as a paddle juggler \cite{sternad2001bouncing}, the impact surface can move as a function determined by time, in which case the guard and reset would depend on the prescribed motion.


To further simplify the component of the saltation matrix \eqref{eq:saltationmatrix} that contains the difference between dynamics, $F_{\mathrm{S}} - \der_xRF_{\mathrm{U}}$, the following steps are applied.
First, substitute in $\dot{q}^+ = M_{\mathrm{S}}^\dagger M \dot{q}^- = \dot{q}^- - J_{\mathrm{S}}^{\dagger T}J_{\mathrm{S}}\dot{q}^-$ using the reset map \eqref{eq::plasticresetmap} and the identity \eqref{eq:mdaggeridentity}.
Then, plugging into the difference between dynamics:
 \begin{align}
     &F_{\mathrm{S}}(t,x^+) - \der_xR_{(\mathrm{U},\mathrm{S})}(t,x^-)F_{\mathrm{U}}(t,x^-) = \label{eq:saltplastic}\\
     &\begin{bmatrix}
     -J_{\mathrm{S}}^{\dagger T}J_{\mathrm{S}}\dot{q}^-\\
    M_{\mathrm{S}}^{\dagger}(C^-\dot{q}^--C^+\dot{q}^+) - J_{\mathrm{S}}^{\dagger T} \dot{J}_{\mathrm{S}}^+\dot{q}^+ - \der_q(M_{\mathrm{S}}^\dagger M\dot{q}^-)\dot{q}^-
     \end{bmatrix}
     \nonumber
 \end{align}
The saltation matrix for plastic impact is obtained by inserting all terms into \revise{the definition of the saltation matrix }\eqref{eq:saltationmatrix} and simplifying (using \eqref{eq:mdaggeridentity} again):
\begin{empheq}[box=\fbox]{align}
    \Xi_{(\mathrm{U},\mathrm{S})}&= \begin{bmatrix}
    M_{\mathrm{S}}^\dagger M & 0_{m\times m}\\
    Z_{\mathrm{S}}+\der_q(M_{\mathrm{S}}^{\dagger}M \dot{q}^-) & M_{\mathrm{S}}^\dagger M
    \end{bmatrix}
    \label{eq:saltplasticsimplify}
\end{empheq}
where
\begin{dmath}
    Z_{\mathrm{S}} {:=} \left(M_{\mathrm{S}}^{\dagger}(C^-\dot{q}^- -C^+\dot{q}^+) - J_{\mathrm{S}}^{\dagger T} \dot{J}_{\mathrm{S}}^+ \dot{q}^+ - \der_q(M_{\mathrm{S}}^\dagger M\dot{q}^-)\dot{q}^-\right ) J_{\mathrm{S}}/(J_{\mathrm{S}}\dot{q}^-)
    \label{eq::saltdifference1}
\end{dmath}
Note that the difference from the Jacobian of the reset map $\der_xR$, \eqref{eq:resetmapjacobianplastic}, is in the first \revise{block} column of the matrix where the identity matrix is now $M_{\mathrm{S}}^\dagger M$ and the element on the lower left differs by the term in \eqref{eq::saltdifference1}.

When impacting into the frictional constrained mode $\mathrm{C}$, all steps remain the same except with $J_{\mathrm{C}}$ instead of $J_{\mathrm{S}}$ (and similarly $M^\dagger_{\mathrm{C}}$ and $J^{\dagger T}_{\mathrm{C}}$).
However, the upper left block of the saltation matrix no longer simplifies as nicely with the Jacobian of the guard $\der_x g$ terms.
This is because $J_{\mathrm{S}} = \der_x g = J_{\mathrm{n}}$ but $J_{\mathrm{C}} \neq \der_x g$.
Rather, $\der_x g = J_{\mathrm{n}}$ is a row of $J_{\mathrm{C}}$, i.e.\ the non-penetrating constraint.
The resulting saltation matrix is
\begin{empheq}[box=\fbox]{align}
    \Xi_{(\mathrm{U},\mathrm{C})}&= \begin{bmatrix}
    \idm-\frac{J_{\mathrm{C}}^{\dagger T}J_{\mathrm{C}}\dot{q}^- J_{\mathrm{n}}}{J_{\mathrm{n}}\dot{q}^-} & 0_{m\times m}\\
    Z_{\mathrm{C}}+\der_q(M_{\mathrm{C}}^{\dagger}M \dot{q}^-) & M_{\mathrm{C}}^\dagger M
    \end{bmatrix}
    \label{eq:salt_u_c}
\end{empheq}
where
\begin{dmath}
    Z_{\mathrm{C}} {:=} \left(M_{\mathrm{C}}^{\dagger}(C^-\dot{q}^- -C^+\dot{q}^+) - J_{\mathrm{C}}^{\dagger T} \dot{J}_{\mathrm{C}}^+ \dot{q}^+ - \der_q(M_{\mathrm{C}}^\dagger M\dot{q}^-)\dot{q}^-\right )J_{\mathrm{C}}/(J_{\mathrm{C}}\dot{q}^-)
    \label{eq::saltdifferenceconstrained}
\end{dmath}
Again, the difference between the saltation matrix and the Jacobian of the reset is in the left column associated with the configuration variations.
However, the upper left block no longer maps configuration variations exactly the same as velocity variations in the lower right, because the tangential constraint is only a velocity constraint -- the contact point can be anywhere on the contact surface, whereas the velocity of the contact point must be the same everywhere on the surface.

Other than the upper left block, the structure of $(\mathrm{U},\mathrm{S})$ and $(\mathrm{U},\mathrm{C})$ saltation matrices look remarkably similar, with the interchange of $J_{\mathrm{S}}$ and $J_{\mathrm{C}}$ being the only other difference. 
In the example in Sec. \ref{sec:salt_example}, the lower left block of these saltation matrices was zero.
This block is comprised of Coriolis-like terms, so for simple systems like the ball drop, Coriolis terms do not exist in the dynamics and the lower left block of the saltation matrix collapses to zero.
However, for systems of appreciable complexity, this does not hold.

\subsection{Elastic impact}
\label{sec:elastic}
When the coefficient of restitution is non-zero, states in the approaching unconstrained mode $\mathrm{U}$ transition directly to the separating unconstrained mode $\mathrm{V}$ through elastic impact.
The dynamics for each mode, \eqref{eq::dynamics}, are
\begin{align}
    F_{\mathrm{U}}(t,x^-) &= \begin{bmatrix}
    \dot{q}\\
    M^{-1}(\Upsilon - C^-\dot{q}^--N)
    \end{bmatrix}\\
    F_{\mathrm{V}}(t,x^+) &= \begin{bmatrix}
    \dot{q}^+\\
    M^{-1}(\Upsilon - C^+\dot{q}^+-N)
    \end{bmatrix}
\end{align}
Again, note that $-$ or $+$ on $\mathrm{C}$ indicates that these functions use the pre- or post-impact velocity, $\dot{q}^-$ or $\dot{q}^+$, respectively.
The Jacobian of the reset map for elastic impact, \eqref{eq::elasticresetmap}, is
\begin{align}
    \der_xR_{({\mathrm{U,V}})}^- = \begin{bmatrix}
    \idm & 0_{m\times m}\\
    \!\der_q(\!(M^\dagger_{\mathrm{n}} M - eJ^{\dagger T}_{\mathrm{n}}J_{\mathrm{n}}) \dot{q}^-) & M^\dagger_{\mathrm{n}} M - eJ^{\dagger T}_{\mathrm{n}}J_{\mathrm{n}}\!
    \end{bmatrix} 
    \label{eq:resetmapjacobianelastic}
\end{align}
The Jacobian of the guard is again $\der_xg = [J_{\mathrm{n}} \ 0_{1\times n}]$.
Plugging each component back into the full saltation matrix equation results in

\begin{empheq}[box=\fbox]{align}
    \Xi_{(\mathrm{U,V})}\! &=\! \begin{bmatrix}
    M^\dagger M - eJ^{\dagger T}J & 0_{m\times m}\\
    \!Z_{\mathrm{V}}\!+\!\der_q(\!(M^\dagger M - eJ^{\dagger T}J)\dot{q}^-)  & \!\!\!M^\dagger M\! -\! eJ^{\dagger T}\!J\!
    \end{bmatrix}
    \label{eq:saltelasticsimplify}
\end{empheq}
where $J$ and $M^\dagger$ use the normal constraint, $J_{\mathrm{n}}$ and $M^\dagger_{\mathrm{n}}$, and
\begin{dmath}
    Z_{\mathrm{V}} {:=} \left(\left[M^{-1}(C^--C^+(M^{\dagger}_{\mathrm{n}} M-eJ^{\dagger T}_{\mathrm{n}} J_{\mathrm{n}})) -\der_q((M^\dagger_{\mathrm{n}} M - eJ^{\dagger T}_{\mathrm{n}} J_{\mathrm{n}}) \dot{q}^-)\right]\dot{q}^- +(1+ e)J^{\dagger T}_{\mathrm{n}} J_{\mathrm{n}} M^{-1}(\Upsilon - C^-\dot{q}^- -N)\right) J_{\mathrm{n}}/(J_{\mathrm{n}}\dot{q}^-)
    \label{eq::saltdifference2}
\end{dmath}
Note that the following substitution can be made $M^\dagger M - eJ^{\dagger T}J = \idm-(1+e)J^{\dagger T}J$ by \eqref{eq:mdaggeridentity}.


\subsection{Stick-slip friction}
\label{sec:friction}
\revise{Stick-slip friction refers to when the tangential force of an object constrained to a surface exceeds the frictional force which constrains the tangential direction. Coulomb friction is a commonly used model of friction in robotics and the experimental law states that the magnitude of the frictional force is equal to the product between the coefficient of static friction $\mu$ and the magnitude of the normal force $f_{\mathrm{n}}(t,x)$.} The saltation matrix for stick-slip friction has been calculated in \cite[Sec.~7.3]{leine2004dynamics}.
This section computes this saltation matrix for a generalized system and analyzes its components. 

When the friction cone is broken, the mode is switched from the constrained mode $\mathrm{C}$ to the sliding mode $\mathrm{S}$.
The guard to check for slipping is \revise{when the tangential force $f_t(t,x)$ exceeds the frictional force $\mu_{s}f_{\mathrm{n}}(t,x)$ in either direction}:
\begin{equation}
    g_{(\mathrm{C,S})}(t,x) := \mu_sf_{\mathrm{n}}(t,x) - f_{\mathrm{t}}(t,x) =0 
\end{equation}
where $\mu_s$ is the coefficient of static friction.
The reset map for these hybrid transitions is an identity transformation $x^+ = R_{(\mathrm{C,S})}(x^-)= x^-$, and therefore $\der_xR_{(\mathrm{C,S})}=\id$.

If the guard $g_{(\mathrm{C,S})}$ is met, it can be assumed that slipping will also occur in the direction of the maximum tangential force.
Therefore, at the slipping boundary, if both the coefficient of static friction and kinetic friction match, $\mu_s=\mu_k$, then $\Delta F = 0$ (as the frictional force reaches and then maintains the value in \eqref{eq:coulomb}) and the saltation matrix is identity by \eqref{eq:identitysaltation}.
Indeed, any friction model (not just Coulomb) where the frictional force matches at the boundary results in an identity saltation matrix.
This includes models where $\mu_k$ is a function of velocity, such as Stribeck friction, so long as at $\|v_{\mathrm{t}}\| = \|J_{\mathrm{t}}\dot{q}\| = 0$, $\mu_k(0) = \mu_s$, to get
 \revise{\begin{empheq}[box=\fbox]{align}
 \mu_s=\mu_k\implies& F_{\mathrm{S}}=F_{\mathrm{C}}
     \implies \Xi_{(\mathrm{C,S})} = \id 
     \label{eq:salt_c_s_equal}
 \end{empheq}}
If $\mu_s \neq \mu_k$, the saltation matrix is not necessarily identity, and the general computations of the saltation matrix can be made to obtain this form:
 \begin{empheq}[box=\fbox]{align}
     \mu_s\neq\mu_k\implies& F_{\mathrm{S}}^+ \neq F_{\mathrm{C}}^- \nonumber\\
     \implies \Xi_{(\mathrm{C,S})} &= \id + \frac{(F_{\mathrm{S}}^+-F_{\mathrm{C}}^-) \der_xg_{(\mathrm{C,S})}}{\der_tg_{(\mathrm{C,S})}+\der_xg_{(\mathrm{C,S})} F_{\mathrm{C}}^-}
     \label{eq:salt_c_s_unequal}
 \end{empheq}
For this saltation matrix, position variations do not change because the reset map is identity and the top row of $F_{\mathrm{S}}$ and $F_{\mathrm{C}}$ are equal (i.e.\ the velocity $\dot{q}$ does not change between modes).

However, this saltation matrix will be very prone to modeling errors as it depends on knowing exactly how the sliding and sticking coefficients differ.
\revise{Because there are many different types of friction models, it may be advantageous to assume that at the boundaries the sliding and sticking coefficients match when appropriate for the specific friction interaction i.e. selecting a Stribeck friction model}.

\subsection{Slip-stick friction}
When the tangential velocity in mode $\mathrm{S}$ goes to zero, the sliding stops and ``sticks'' into the constrained mode $\mathrm{C}$.
Therefore, the guard at slip-stick friction is just the magnitude of the tangential velocity:
\begin{align}
    g_{(\mathrm{S,C})}(t,x) &:=\|J_{\mathrm{t}}\dot{q}\|= \|v_{\mathrm{t}}\|\\
    \der_xg_{(\mathrm{S,C})}(t,x) &= \begin{bmatrix} \dot{J}_{\mathrm{t}}^-  & J_{\mathrm{t}} \end{bmatrix}
\end{align}
The guard also has the condition $f_{\mathrm{t}} < \mu_s f_{\mathrm{n}}$. However, note that the way tangential friction forces are calculated is different in the sliding mode $\mathrm{S}$ than in the sticking mode $\mathrm{C}$. In sliding, the tangential force is proportional to the normal force, $f_{\mathrm{t}} = \mu_k f_{\mathrm{n}}$, \eqref{eq:coulomb}. In the constrained sticking mode, the force vector is calculated from Lagrange multipliers as in \eqref{eq:fdynamics}. These generally are not equal and so there is a difference in the tangential force at the transition, and thus a difference in dynamics.

The reset is an identity transformation, $x^+ = R_{(\mathrm{S,C})}(x^-)= x^-$, and therefore $\der_xR_{(\mathrm{S,C})}=\id$, so the saltation matrix is primarily composed of the difference between the dynamics of both modes and the tangential velocity term from the guard.
\revise{The guard captures time varying interactions with the environment through the constraint Jacobian's $J(t,q)$ derivatives. 
There are no other time varying components which enter so the simplification $\der_t g_{(\mathrm{S,C})} = 0$ can be made}. The saltation matrix is
\begin{empheq}[box=\fbox]{align}
    \Xi_{(\mathrm{S,C})} = \id + \frac{(F_{\mathrm{C}}^+-F_{\mathrm{S}}^-) \begin{bmatrix} \dot{J}_{\mathrm{t}}^-  & J_{\mathrm{t}} \end{bmatrix}}{ \dot{J}_{\mathrm{t}}^- \dot{q}^- + J_{\mathrm{t}} \ddot{q}^-}
    \label{eq:salt_s_c}
\end{empheq}
Note that the denominator is the tangential acceleration constraint \eqref{eq:accelcons} in mode $\mathrm{C}$.
If this condition is met at the exact moment that the velocity guard is satisfied while in the sliding mode $\mathrm{S}$, the saltation matrix is not well defined;
however, this would violate the transversality assumption \eqref{asm:transverse}.
For this saltation matrix, as with stick-slip, position variations do not change because the reset map is identity and the top row of $F_{\mathrm{C}}$ and $F_{\mathrm{S}}$ are equal (i.e.\ the velocity $\dot{q}$ does not change between modes).

\subsection{Analysis of Saltation Matrices for Rigid Bodies}

This section presents saltation matrix derivations for a number of hybrid transitions that occur in rigid body systems with contact, as summarized in Table~\ref{table:saltation_structure}.
These derivations reveal patterns among many of these saltation matrices.
For instance, the upper right block of the saltation matrix is zero for every case presented here.
This is due to the second order nature of mechanical systems as a whole (i.e.\ acceleration is the derivative of velocity, which is the derivative of position).
This makes it convenient to perform the eigen-analysis as in Sec.~\ref{sec:salt_example}.
The eigenvalues and eigenvectors of a block triangular matrix are the eigenvalues and eigenvectors of its diagonal block components, and the lower left block does not affect them.
In applications where only the eigenvalues of the saltation matrix of interest, knowing the structure of the saltation matrix means the full saltation matrix need not be computed.

Four of the saltation matrices analyzed are identity: apex \eqref{eq:salt_v_u}, the two liftoff cases (\ref{eq:salt_c_v},\ref{eq:salt_s_v}), and stick-slip under constant friction coefficient \eqref{eq:salt_c_s_equal}.
This occurs when the reset map is identity and the dynamics in each mode are equivalent, as in \eqref{eq:identitysaltation}.
Outside of these identity cases, the stick-slip with unequal friction coefficients \eqref{eq:salt_c_s_unequal} and slip-stick \eqref{eq:salt_s_c} transitions also have an identity reset map because there is no instantaneous change in positions or velocities.
An identity reset map allows for further insight into the eigen-properties of these matrices.
Both of these saltation matrices can be written as $\Xi = I_{n\times n} + ab^T$ where $a$ and $b$ are $n\times 1$ vectors and $ab^T$ is their outer product.
The eigenvalues of a matrix with this structure are all $1$ except for one eigenvalue of $1+a^Tb$ with corresponding eigenvector $a$.
This can be easily shown from the equality:
\begin{align}
    (I+ab^T)a &=  a + a(b^Ta) = (1+a^Tb)a
\end{align}
This makes it possible to compute the eignvalues of these saltation matrices without performing the full matrix computation.

Two non-identity saltation matrices had equivalent diagonal blocks, sliding plastic impact \eqref{eq:saltplasticsimplify} and elastic impact \eqref{eq:saltelasticsimplify}.
This occurs because the guard surface enforces an equivalent constraint on both position and velocities to be along the guard.
When a non-holonomic constraint is added in mode $\mathrm{C}$, this equivalency breaks.
Equal diagonal blocks means that the eigenvalues of these saltation matrices are the eigenvalues of a diagonal block repeated twice.
Table \ref{table:saltation_structure} summarizes the properties of identity reset map, matching hybrid dynamics, equal diagonal blocks, as well as equation number for each saltation matrix.

\begin{table}[t]
     
    \caption{Properties of the Saltation \revise{Matrix} for Different Rigid Body Mode Transitions}
    \label{table:saltation_structure}
    \centering
    \begin{tabular}{ ccccc } 

    Transition & $R = I$ & $F^+ = F^-$ & Equal Diag. Blocks & Eq. \#\\
     \hline
     $(\mathrm{V},\mathrm{U})$ & \cmark & \cmark & \cmark & \eqref{eq:salt_v_u} \\ 
      $(\mathrm{U},\mathrm{S})$ & \xmark & \xmark & \cmark & \eqref{eq:saltplasticsimplify} \\ 
     
     $(\mathrm{U},\mathrm{C})$ & \xmark & \xmark & \xmark & \eqref{eq:salt_u_c} \\
    
     $(\mathrm{U},\mathrm{V})$ & \xmark & \xmark & \cmark & \eqref{eq:saltelasticsimplify} \\
      $(\mathrm{C},\mathrm{S})$ & \cmark & if $\mu_s = \mu_k$ & if $\mu_s = \mu_k$ & (\ref{eq:salt_c_s_equal},\ref{eq:salt_c_s_unequal}) \\
     $(\mathrm{C},\mathrm{V})$ & \cmark & \cmark & \cmark & \eqref{eq:salt_c_v} \\
     $(\mathrm{S},\mathrm{V})$ & \cmark & \cmark & \cmark & \eqref{eq:salt_s_v} \\ 
      $(\mathrm{S},\mathrm{C})$ & \cmark & \xmark & \xmark & \eqref{eq:salt_s_c}
     \end{tabular}
\end{table}

\section{Conclusion}
The saltation matrix is an essential tool when dealing with hybrid systems with state dependent switches.
This paper presents a derivation of the saltation matrix with two different methods and demonstrates how the saltation matrix can be used in linear and quadratic forms for hybrid systems. 
A survey of where saltation matrices are used in other fields is also presented.
In the past, it has been heavily utilized for analyzing the stability of periodic systems, but more recently it has been critical for analyzing and designing non-periodic behaviors.
This analysis is especially useful for robotics where many important robotic motions are not periodic, but are hybrid due to the discontinuous nature of impact in rigid body systems with unilateral constraints.

To further explore the nature of contact and how variations are mapped through them, a simple contact system is considered to compute the saltation matrix for plastic impact and analyze the different components of the resulting saltation matrices.
These saltation matrices capture how position variations are mapped through contact, whereas the Jacobian of the reset map does not provide any information on position.
In addition to this simple example, saltation matrices are computed for each of the hybrid transitions for a generalized rigid body model and we give insights on their structure.
These computations are especially useful because the rigid body model covers a wide variety of systems and will help when getting started using saltation matrices for these systems.
Saltation matrices exhibit common structures that can be exploited. 
\revise{In contrast to using the Jacobian of the reset map, the saltation matrix captures the position variational information when applying a unilateral holonomic constraint.}
For other hybrid transitions such as stick-slip friction, the Jacobian of the reset map provides no additional information because it is an identity transformation and all the information is contained in the saltation matrix.


\revise{In general, the effects of hybrid transitions are significant and can not simply be ignored. 
But, in specific instances, it's possible to simplify computations by not computing the saltation matrix: when the saltation matrix is identity \eqref{eq:identitysaltation} or when the hybrid transition is time dependent rather than state dependent the Jacobian of the reset map can be used instead \eqref{eq:timetransitionsaltation}.  
In rigid body dynamics with unilateral constraints, for instance, the saltation matrix becomes an identity matrix when a constraint is removed (liftoff), allowing it to be disregarded. However, it is essential to first verify that the saltation matrix is always an identity matrix in such scenarios.
Another example involves systems with slow and fast dynamics. 
If the fast dynamics are hybrid and stable, and the focus is on the slow dynamics, the saltation matrix can be ignored. 
Take a DC motor controlled by pulse-width modulation (PWM) as an example. 
The PWM operates as a fast hybrid system, but the position and velocity of the rotor, which are slow dynamics, are of primary concern.
Knowing that the PWM is stable allows us to ignore the saltation matrices and focus on the average effect of the fast hybrid dynamics (nominal voltage) on the slow dynamics (rotor velocity).
However, once again, the saltation matrices might be needed to show stability of the fast hybrid dynamics before it can be ignored.}

By using saltation matrices for hybrid systems, efficient analysis, planning, control, and state estimation algorithms can be produced.
This is especially important as many algorithms for hybrid systems naturally have combinatoric time complexities and through the use of these tools we can simplify these problems.
The hope of this paper is to introduce the topic of saltation matrices to a broader community so that we can, as a whole, develop better methods for dealing with the complexities of hybrid systems and their applications.


\section*{Acknowledgement}
We would like to thank Professor Sam Burden for his contributions in the conceptualization of this work and for his feedback on early drafts.
We would also like to thank Dr. George Council for his comments and suggestions for additional material to cover.

\appendix
Appendices \ref{appendix:chainrule} and \ref{appendix:salt_derivation_early} present the chain rule derivation of the saltation matrix and the early impact case for the geometric derivation.
Appendices \ref{appendix:covariance} and \ref{appendix:Riccati} prove the update laws through hybrid events for both covariance propagation and the Riccati equations.

\subsection{Saltation matrix chain rule derivation}
\label{appendix:chainrule}
\revise{This appendix shows an alternate derivation of \eqref{eq:saltationmatrix} using the chain rule, providing an analytical derivation rather than a geometric one.}
Define the solutions of the flow in hybrid domains I and J, which integrate the continuous dynamics from an initial state $x$ at time $t_0$ to a state $x_f$ at time $t_f$, as
\begin{align}
    \phi_{\mathrm{I}}: (t_0 \in \mathbb{R}, t_f\in\mathbb{R},x\in{D}_{\mathrm{I}}) \mapsto x_f\in{D}_{\mathrm{I}} \label{eq:phii}\\
    \phi_{\mathrm{J}}: (t_0 \in \mathbb{R},t_f\in\mathbb{R},x\in{D}_{\mathrm{J}}) \mapsto x_f\in{D}_{\mathrm{J}}
\end{align}
such that the vector fields \revise{are}
\begin{align}
    F_{\mathrm{I}}(t_0,x) &= -\der_{t_0} \phi_{\mathrm{I}}(t_0,t_f,x) \label{eq:dt0phi}\\ 
    F_{\mathrm{I}}(t_f,x) &= \der_{t_f} \phi_{\mathrm{I}}(t_0,t_f,x) \label{eq:dtfphi}
\end{align}
for each mode.
Define the solution across a hybrid transition from mode I to J to be
 \begin{align}
    \phi(t_0,t_f,x) := &\phi_{\mathrm{J}}(\tau(x),t_f,R_{(\mathrm{I},\mathrm{J})}(\tau(x),\phi_{\mathrm{I}}(t_0,\tau(x),x)))\label{eq:combined_flow}
\end{align}
where $\tau(x)$ is the time to impact map, such that
\begin{align}
    g_{(\mathrm{I},\mathrm{J})}(\tau(x),\phi_{\mathrm{I}}(t_0,\tau(x),x))&=0 \label{eq:tau_x_def}
\end{align}
It helps to look at the in between steps of the function composition in \eqref{eq:combined_flow}. Define
\begin{align}
    x^-(x) &:= \phi_{\mathrm{I}}(t_0,\tau(x),x) \label{eq:x_minus}\\
    x^+(x) &:= R_{(\mathrm{I},\mathrm{J})}(\tau(x),x^-(x))\\
    x_f(x) &:= \phi_{\mathrm{J}}(\tau(x),t_f,x^+(x))\label{eq:x_f}
\end{align}
where $x_f=\phi(t_0,t_f,x)$ is the final state in the new mode.
To find the derivative of $\phi$ with respect to $x$ in \eqref{eq:combined_flow}, the chain rule is used on each of these steps:
\begin{align}
    \der_x x^-(x) &= \der_{\tau(x)} \phi_{\mathrm{I}} \der_x \tau + \der_x \phi_{\mathrm{I}} \label{eq:dx_x_minus}\\
    \der_x x^+(x) &= \der_{\tau(x)} R_{(\mathrm{I},\mathrm{J})} \der_x \tau + \der_{x^-(x)} R_{(\mathrm{I},\mathrm{J})} \der_x x^-\\
    \der_x x_f(x) &= \der_{\tau(x)} \phi_{\mathrm{J}}  \der_x \tau + \der_{x^+(x)} \phi_{\mathrm{J}} \der_x x^+
\end{align}
where the arguments to each function are suppressed but equal to their corresponding value in \eqref{eq:x_minus}--\eqref{eq:x_f}.

Combining these:
\begin{dmath}
    \der_x \phi =
    \der_{\tau(x)} \phi_{\mathrm{J}}  \der_x \tau + \der_{x^+(x)} \phi_{\mathrm{J}} [\der_{\tau(x)} R_{(\mathrm{I},\mathrm{J})} \der_x \tau + \der_{x^-(x)} R_{(\mathrm{I},\mathrm{J})} ( \der_{\tau(x)} \phi_{\mathrm{I}} \der_x \tau + \der_x \phi_{\mathrm{I}})]
    \label{eq:chainrule_first}
\end{dmath}


As this is a first order approximation, the terms $\der_{x} \phi_{\mathrm{I}}$ and $\der_{x^+(x)} \phi_{\mathrm{J}}$ can be taken as identity matrices (as they would in a linear system), and so this simplifies to (with additional substitutions for $F_{\mathrm{I}}$ and $F_{\mathrm{J}}$ using \eqref{eq:dt0phi}--\eqref{eq:dtfphi}):
\begin{dmath}
    \der_x \phi =
    (- F_{\mathrm{J}}   + \der_{\tau(x)} R_{(\mathrm{I},\mathrm{J})} + \der_{x^-(x)} R_{(\mathrm{I},\mathrm{J})} F_{\mathrm{I}}) \der_x \tau + \der_{x^-(x)} R_{(\mathrm{I},\mathrm{J})}
    \label{eq:chainrule_second}
\end{dmath}
To obtain $\der_x\tau$, use the implicit function theorem and take the chain rule on the guard condition \eqref{eq:tau_x_def}, and using \eqref{eq:dtfphi} and \eqref{eq:dx_x_minus} results in the following relation:
\begin{align}
    &0= \der_{\tau(x)}g_{(\mathrm{I},\mathrm{J})} \der_x \tau(x) + \der_{x^-(x)}g_{(\mathrm{I},\mathrm{J})}\der_x x^-\\
    &0=\left(\der_{\tau(x)}g_{(\mathrm{I},\mathrm{J})} + \der_{x^-(x)}g_{(\mathrm{I},\mathrm{J})}F_{\mathrm{I}} \right)\der_x \tau + \der_{x^-(x)}g_{(\mathrm{I},\mathrm{J})} \\
    &\der_x \tau(x) = \frac{- \der_{x^-(x)} g_{(\mathrm{I},\mathrm{J})}}{\der_{\tau(x)} g_{(\mathrm{I},\mathrm{J})} + \der_{x^-(x)} g_{(\mathrm{I},\mathrm{J})} F_{\mathrm{I}}}
\end{align}

Plugging back into \eqref{eq:chainrule_second},
evaluating at the instant of impact, $t=\tau(x)=0$,
substituting the notation from \eqref{eq:F_I_minus}--\eqref{eq:D_t_g}, and simplifying:
\begin{align}
    \der_x \phi &= \der_xR^- + \left(-F_{\mathrm{J}}^+ + \der_tR^- + \der_xR^-F_{\mathrm{I}}^-  \right)\der_{x}\tau\\
        \der_x \phi &= \der_xR^- + \frac{\left(F_{\mathrm{J}}^+-\der_xR^-F_{\mathrm{I}}^- - \der_tR^- \right) \der_x g^-}{\der_t g^- + \der_x g^- F_{\mathrm{I}}^-}\\
    \der_x \phi &:= \Xi_{(\mathrm{I},\mathrm{J})}
\end{align}
as in \eqref{eq:saltationmatrix}, where all terms are evaluated at the time of impact and the state just before impact, except for $F_{\mathrm{J}}^+$ which is evaluated at the state just after impact, as in \eqref{eq:F_I_minus}--\eqref{eq:D_t_g}.

\subsection{Early impact saltation derivation}
\label{appendix:salt_derivation_early}
In the geometric derivation of the saltation matrix, it was assumed the perturbed trajectory impacted late. This appendix shows that the saltation matrix expression is the same if derived following the same logic as Sec.~\ref{sec:derivation} but with early impact.
It may help to visualize Fig. \ref{fig:geometric-saltation} with the roles of the nominal $x(t)$ and perturbed $\pert{x}(t)$, and the corresponding linearization arrows, flipped.

Again, start by assuming the same flow, reset, and guard linearizations as in 
\eqref{eq:constant_flow}--\eqref{eq:guard_linearization}. 
The perturbed impact occurs first at time $\pert{t}^-$ i.e.\ $\pert{t}^-<t^-$ and $\delta t = \pert{t}^- - t^- < 0$.
Because the perturbed trajectory impacts first, the aim is to find the mapping from $\delta x(\pert{t}^-)$ to $\delta x(t^+)$ (instead of $\delta x({t}^-)$ to $\delta x(\pert{t}^+)$ as in the case of late impact).
This allows for comparisons between states (nominal and perturbed) that are in the same hybrid domain.

Define $\delta x(\pert{t}^-)$ and $\delta x({t}^+)$ to be
\begin{align}
   \delta x(\pert{t}^-) &:= \pert{x}(\pert{t}^-) - x(\pert{t}^-)\label{eq:early_impact_delta_pre}\\
    \delta x(t^+) &:= \pert{x}({t}^+) - x({t}^+)
    \label{eq:early_impact_delta_post}
\end{align}
We would like to write these in terms of the nominal trajectory at that time.
Using the linearization of the flow before impact \eqref{eq:constant_flow} and rearranging  \eqref{eq:early_impact_delta_pre} we get 
\begin{align}
    \pert{x}(\pert{t}^-) = x(t^-) + \delta x(\pert{t}^-) + F_{\mathrm{I}}^- \delta t
    \label{eq:early_impact_pert_x_minus_x_minus}
\end{align}
Since the perturbed trajectory impacts earlier, next is to compute where it ends up after the reset map is applied and it flows for $|\delta t|$ time on the new dynamics.
Again, using the linearization of the flow \eqref{eq:constant_flow_J}:
\begin{align}
    \pert{x}(t^+) &= R(\pert{t}^-,\pert{x}(\pert{t}^-)) - F_{\mathrm{J}}^+ \delta t
    \label{eq:early_impact_x_pert_plus}
\end{align}
Next, $R(\pert{t}^-,\pert{x}(\pert{t}^-))$ can be solved for as a function of $x(\pert{t}^-)$
by substituting in $\pert{x}(\pert{t}^-)$ from \eqref{eq:early_impact_pert_x_minus_x_minus} and using the linearization of the reset map from \eqref{eq:reset_linearization}:
\begin{align}
    \bar{R}(\pert{t}^-,\pert{x}(\pert{t}^-)) &= \bar{R}(t^- + \delta{t},x(t^-)  + \delta x(\pert{t}^-) + F_{\mathrm{I}}^- \delta t)\\
    = R(t^-,x(t^-))&+\der_xR^-\left(\delta x(\pert{t}^-)+ F_{\mathrm{I}}^- \delta t\right) + \der_t{\mathrm{R}}^-\delta{t}
\end{align}
Now plugging back in to \eqref{eq:early_impact_x_pert_plus}:
\begin{align}
    \pert{x}(t^+) =& R(t^-,x(t^-))+\der_xR^-\delta x(\pert{t}^-) \\ &+\left(\der_xR^-F_{\mathrm{I}}^- + \der_t{\mathrm{R}}^- - F_{\mathrm{J}}^+\right) \delta t \nonumber
\end{align}
$\delta x(t^+)$ can be written as a function of $\delta x(\pert{t}^-)$ and $\delta t$ by subbing $\pert{x}(t^+)$ into \eqref{eq:early_impact_delta_post}:
\begin{align}
    \delta x(t^+) &= \der_xR^-\delta x(\pert{t}^-) +\left(\der_xR^-F_{\mathrm{I}}^- + \der_t{\mathrm{R}}^- - F_{\mathrm{J}}^+\right) \delta t
    \label{eq:early_impact_almost_there}
\end{align}

Next, $\delta t$ can be found as a function of $\delta x(\pert{t}^-)$ using
\begin{align}
    0 &= g(\pert{t}^-,\pert{x}(\pert{t}^-))
\end{align}
Substituting in \eqref{eq:early_impact_pert_x_minus_x_minus} and expanding using the linearization of the guard \eqref{eq:guard_linearization} (and noting that $g(t^-,x(t^-)=0$):
\begin{align}
    0 &= g\left(t^- + \delta{t},x(t^-) + \delta x(\pert{t}^-) + F_{\mathrm{I}}^- \delta t\right)\\
    0 &= g(t^-,x(t^-)) + \der_xg^-\left( \delta x(\pert{t}^-)+F_{\mathrm{I}}^- \delta t\right) + \der_tg^-\delta{t}\\
        0 &= \der_xg^-\left( \delta x(\pert{t}^-) +F_{\mathrm{I}}^- \delta t \right)+ \der_tg^-\delta{t}
\end{align}
Now, solving for $\delta t$ in terms of $\delta x (\pert{t}^-)$:
\begin{align}
    \delta t &= -\frac{\der_xg^-}{\der_x g^- F_{\mathrm{I}}^-+ \der_tg^-} \delta x(\pert{t}^-)
    \label{eq:early_impact_delta_t}
\end{align}

Substitute \eqref{eq:early_impact_delta_t} into \eqref{eq:early_impact_almost_there} and simplify to get the saltation matrix equivalent to \eqref{eq:saltsensitivity}:
\begin{align}
    \delta x(t^+) &= \der_xR^-\delta x(\pert{t}^-) \\
    & \quad +\frac{\left(F^+_{\mathrm{J}} - \der_xR^- F^-_{\mathrm{I}} - \der_tR^-\right)\der_x g^- }{\der_x g^- F^-_{\mathrm{I}} +\der_t g^-}\delta x(\pert{t}^-)\nonumber\\
   & = \Xi_{(\mathrm{I},\mathrm{J})} \delta x(\pert{t}^-) \label{eq:saltsensitivit_new}
\end{align}

\subsection{Covariance update through a hybrid event}
\label{appendix:covariance}
This appendix presents a derivation for the covariance update through a reset map, \eqref{eq:covarianceupdate}.
Consider the state trajectory as a random variable $X(t)$ with mean $\rho(t)=x(t)$, the nominal trajectory $x(t)$, and covariance $\Sigma(t)$. Define a perturbation as a zero mean random variable $\delta x(t)$ with the same covariance, such that $X(t) = x(t) + \delta x(t)$, \revise{where $X(t)$, $x(t)$, and $\delta x(t)$ evolve according to the dynamics of the hybrid system. Therefore, once $X(0)$ and $\delta x(0)$ are sampled, the dynamics evolve deterministically.} 

At a hybrid impact event, define the pre-impact time of the mean to be $t^-$, where $g(t^-,\rho(t^-)) = 0$, and the corresponding post-impact time to be $t^+$. Consider how the distribution is updated to find $X(t^+)$ based on $X(t^-)$.
To find the mean, take the expectation of $X(t^+)$:
\begin{align}
    \rho(t^+) = &\mathbb{E}[X(t^+)] = \mathbb{E}[x(t^+)+\delta x(t^+)]\\
     =& x(t^+)+\mathbb{E}[\delta x(t^+)]
\end{align}
where the two terms are separable because expectation is a linear operator,
and the expectation of the nominal post-impact state is just its value, $\mathbb{E}[x(t^+)]=x(t^+)=R(x(t^-))$. 
Substituting in $\delta x(t^+) = \Xi_{(\mathrm{I},\mathrm{J})}(t^-,x(t^-))\delta x(t^-) + \hot$ from \eqref{eq::saltperturbation}:
\begin{align}
    \rho(t^+) = x(t^+)+\mathbb{E}[\Xi_{(\mathrm{I},\mathrm{J})}(t^-,x(t^-))\delta x(t^-) + \hot]\\
    \rho(t^+) = x(t^+)+\Xi_{(\mathrm{I},\mathrm{J})}(t^-,x(t^-))\mathbb{E}[\delta x(t^-)] + \mathbb{E}[\hot]
\end{align}
Because expectation is a linear operator, $\Xi(t^-,x(t^-))$ can be moved out of the expectation. Then, because $\delta x(t^-)$ is centered about zero, $\mathbb{E}[\delta x(t^-)] = 0$, and for small displacements the higher order terms are negligible, $\mathbb{E}[\hot] \approx 0$, which simplifies to
\begin{equation}
    \rho(t^+) \approx x(t^+) = R(x(t^-))
    \label{eq::resetupdate}
\end{equation}
\revise{Note that it is approximate because of the approximation that higher order terms are zero.}

Covariance is defined as
\begin{equation}
    \mathbb{COV}[X] := \mathbb{E}[(X-\mathbb{E}[X])(X-\mathbb{E}[X])^T ]
   \end{equation}
the post-impact covariance $\Sigma(t^+)$ is
\begin{align}
    \Sigma(t^+) =& \mathbb{COV}[X(t^+)]
    = \mathbb{COV}[x(t^+)+\delta x(t^+)] \\
    = &\mathbb{E}\Big[\Big((x(t^+)+\delta x(t^+)-\rho(t^+))\nonumber\\
   &\qquad (x(t^+)+\delta x(t^+)-\rho(t^+)\Big)^T \Big]
   \end{align}
Since $\rho(t^+) = x(t^+)$, this simplifies to
\begin{dmath}
   \Sigma(t^+) =\mathbb{E}[\delta x(t^+)\delta x(t^+)^T ]
\end{dmath}
Using \eqref{eq::saltperturbation}, $\delta x(t^+)$ can be expanded as
\begin{align}
    \Sigma(t^+) &=\mathbb{E}[(\revise{\Xi_{(I,J)}}\delta x(t^-)  + \hot)(\revise{\Xi_{(I,J)}}\delta x(t^-)  + \hot)^T ]\\
& =\revise{\Xi_{(I,J)}}\mathbb{E}[\delta x(t^-) \delta x(t^-)^T]\revise{\Xi^T_{(I,J)}}  \\
& + 2\revise{\Xi_{(I,J)}}\mathbb{E}[\delta x(t^-)(\hot)^T]+ \mathbb{E}[(\hot)(\hot)^T]
\end{align}
and for small displacements, $\hot \approx 0$, which simplifies to
\begin{align}
    \Sigma(t^+) \approx \revise{\Xi_{(I,J)}}\Sigma (t^-)\revise{\Xi^T_{(I,J)}}
    \label{eq::saltcovarianceupdate}
\end{align}
as in \eqref{eq:covarianceupdate}, which holds to first order and is exact for linear hybrid systems.

\subsection{Riccati update through hybrid events}
\label{appendix:Riccati}
This appendix derives the update for the Riccati equation through a hybrid event, \eqref{eq::saltriccatiupdate}. \revise{This general form can be used to update co-vectors through hybrid transitions when the co-vector can be represented as a matrix multiplication of a quadratic form and the change in state $p = P\delta x$.}
See \cite[Ch. 6.1]{liberzon2012calculus} for a background on the continuous Riccati update and \cite[Ch. 8.3]{underactuated} for an overview of the discrete formulation.
Solving the Riccati update along a trajectory yields a locally optimal feedback controller, called the linear quadratic regulator (LQR).
The optimality of LQR is conditioned on the balance between penalties on deviations in state $Q$ and control input $V$ at each timestep, called the stage cost, and at the final state, called the terminal cost, where $Q$ is a positive semi-definite matrix and $V$ is positive-definite.

Define the optimal stage cost $\ell_{t^-}^*$ for the reference trajectory $\left(x(t),u(t)\right)$ and the optimal solution $\left({x}^*,u^*\right)$ applied at a hybrid transition at time $t^-$ as
\begin{align}
    \ell_{t^-}^* = \ell_{t^-}&(x^*(t^-),{u}^*(t^-)) =  \nonumber \\ &\frac{1}{2}(x^*(t^-)-x(t^-))^TQ_{t^-}(x^*(t^-)-x(t^-)) \nonumber \\
    +&\frac{1}{2}(u^*(t^-)-u(t^-))^TV_{t^-}(u^*(t^-)-u(t^-))
    \label{eq:opt_cost_to_go}
\end{align}
where $Q_{t^-}$ and $V_{t^-}$ are the quadratic penalty on state and input respectively at time $t^-$.
Define the current state to be $\pert{x}$ and the difference with the optimal solution to be
\begin{align}
    \delta x^*(t^-) := x^*(t^-)-\pert{x}(t^-)
\end{align}
such that \eqref{eq:opt_cost_to_go} becomes
\begin{align}
    &\ell_{t^-}^* = \frac{1}{2}(\delta x^*)^TQ_{t^-}(\delta x^*)\nonumber\\&+\frac{1}{2}(u^*(t^-)-u(t^-))^TV_{t^-}(u^*(t^-)-u(t^-))
\end{align}
Because the transition is instantaneous, assume that the input has no effect $u(t^-) = u^*(t^-)$ and simplify the optimal stage cost as
\begin{align}
    &\ell_{t^-}^* = \frac{1}{2}(\delta x^*)^TQ_{t^-}(\delta x^*)
\end{align}
The Hamiltonian \cite[Ch. 2.4]{liberzon2012calculus} for the hybrid transition \revise{is}
\begin{align}
    H_{t^-}&:= H(x^*(t^-),u^*(t^-),p^*(t^+))\nonumber\\
    &:= \ell_{t^-}^*+ R^{T}_{(\mathrm{I},\mathrm{J})}(t^-,x^*(t^-))p^*(t^+)
    \label{eq:hamiltonian}
\end{align}
where $p^*(t^+)$ is the optimal costate \cite[Ch. 3.4]{liberzon2012calculus}.
Using the expansion \eqref{eq::saltperturbation} about $R_{(\mathrm{I},\mathrm{J})}(t^-,\pert{x}(t^-)+\delta x^*(t^-))$ :
\begin{equation}
    R_{(\mathrm{I},\mathrm{J})}(t^-,x^*(t^-))= R_{(\mathrm{I},\mathrm{J})}(t^-,\pert{x}(t^-)) + \Xi\delta x^*(t^-) + \hot
\end{equation}
where $\Xi = \Xi_{(\mathrm{I},\mathrm{J})}(t^-,\pert{x}(t^-))$.
The Hamiltonian for the hybrid transition is then
\begin{dmath}
    H_{t^-} = \frac{1}{2}(\delta x^*(t^-))^T Q_{t^-} \delta x^*(t^-) + \left(R_{(\mathrm{I},\mathrm{J})}(t^-,\pert{x}(t^-)) + \Xi\delta x^*(t^-) + \hot\right)^Tp^*(t^+)
\end{dmath}
Using Pontryagin's Maximum principle \cite[Ch. 4.1]{liberzon2012calculus}, derive the optimal state update and costate update:
\begin{align}
x^*(t^+) =& \der_{p^*}H_{t^-} = R_{(\mathrm{I},\mathrm{J})}(t^-,\pert{x}(t^-)) + \Xi
\delta x^*(t^-)\\
p^*(t^-) =& \der_{x^*}H_{t^-} = Q_{t^-}\delta x^*+\Xi^Tp^*(t^+) + \hot
\end{align}
Given the standard costate guess of $p(t^+) = P(t^+)\delta x(t^+)$ \cite{underactuated}, we can derive the hybrid update for the matrix $P$, which defines the boundary conditions for the optimal control problem:
\begin{dmath}
    P(t^-)\delta x^*(t^-) = Q_{t^-}\delta x^*(t^-)+\Xi^TP(t^+)\delta x^*(t^+) +\hot
\end{dmath}
Substitute $\delta x^*(t^+) = \Xi \delta x^*(t^-) + \hot$:
\begin{dmath}
    P(t^-)\delta x^*(t^-) = Q_{t^-}\delta x^*(t^-)+\Xi^TP(t^+) (\Xi\delta x^*(t^-) +\hot)  +\hot
\end{dmath}
The update for $P(t^-)$ is recursive and cannot be computed as is.
However, when higher order terms are small, we cancel $\delta x^*(t^-)$ from both sides and write the Bellman update for $P(t^-)$:
\begin{align}
P(t^-)\delta x^*(t^-) \approx& Q_{t^-}\delta x^*(t^-)+\Xi^TP(t^+)\Xi\delta x^*(t^-)\\
P(t^-) \approx& Q_{t^-}+\Xi^TP(t^+)\Xi
\end{align}

\addtolength{\textheight}{-0.5cm}   



\bibliographystyle{IEEEtran}
\bibliography{references}
\end{document}